\documentclass[journal]{IEEEtran}

\usepackage{graphicx} 
\usepackage{subfig}
\usepackage{cite}
\usepackage{multirow}
\usepackage{enumitem}
\usepackage{amsfonts}
\usepackage{amsmath}
\usepackage[ruled,vlined,lined,linesnumbered,boxed]{algorithm2e}
\usepackage{fancyhdr}
\usepackage{url}
\usepackage{color}
\usepackage{./slam_macros}

\graphicspath{ {./images/}}

\begin{document}
\bstctlcite{IEEEexample:BSTcontrol}
%
\title{Good Feature Matching: Towards Accurate, Robust VO/VSLAM with Low Latency}

\author{Yipu~Zhao,
	and~Patricio~A.~Vela,~\IEEEmembership{Member,~IEEE}
\thanks{Y.~Zhao and P.~A.~Vela are with the School of Electrical and Computer Engineering,
and Institute of Robotics and Intelligent Machines,  
Georgia Institute of Technology, Atlanta,GA, 30332 USA. 
e-mails: (yipu.zhao@gatech.edu, pvela@gatech.edu).
This research was partially funded by the National Science 
Foundation (Award \#1816138).}
}

\onecolumn
\pagenumbering{gobble}
{\noindent \copyright 2020 IEEE.  Personal use of this material is permitted.  Permission from IEEE must be obtained for all other uses, in any current or future media, including reprinting/republishing this material for advertising or promotional purposes, creating new collective works, for resale or redistribution to servers or lists, or reuse of any copyrighted component of this work in other works.}
\newline
\newline
\newline
This paper has been accepted for publication in IEEE Transactions on Robotics.
\newline
\noindent Please cite the paper using the following bibtex:
\newline
\newline
@article\{zhao2020tro-gfm,\\
\indent \indent title \indent \indent = \{Good Feature Matching: Towards Accurate, Robust VO/VSLAM with Low Latency\},\\
\indent \indent author \indent = \{Yipu Zhao and Patricio A. Vela\},\\
\indent \indent journal\indent = \{IEEE Transactions on Robotics\},\\
\indent \indent year\indent \indent = \{2020\},\\
\indent \indent doi \indent \indent = \{10.1109/TRO.2020.2964138\}\\
\}

\newpage
\twocolumn
\pagenumbering{arabic}

\maketitle


\begin{abstract}
Analysis of state-of-the-art VO/VSLAM system exposes a gap in balancing
performance (accuracy \& robustness) and efficiency (latency).
Feature-based systems exhibit good performance, yet have higher latency due to
explicit data association; direct \& semidirect systems have lower latency,
but are inapplicable in some target scenarios or exhibit lower accuracy than
feature-based ones.
This paper aims to fill the performance-efficiency gap with an enhancement
applied to feature-based VSLAM.  
We present good feature matching, an active map-to-frame feature
matching method.  Feature matching effort is tied to submatrix selection,
which has combinatorial time complexity and requires choosing a scoring
metric.  Via simulation, the {\em Max-logDet} matrix revealing metric is shown
to perform best.  For real-time applicability, the combination of
deterministic selection and randomized acceleration is studied.  
The proposed algorithm is integrated into monocular \& stereo feature-based
VSLAM systems.  Extensive evaluations on multiple benchmarks and compute
hardware quantify the latency reduction and the accuracy \& robustness
preservation.  
\end{abstract}


\IEEEpeerreviewmaketitle


\section{Introduction} \label{sec::intro}
Pose tracking with vision sensors has application to Robotics and Augmented
Reality (AR). 
Research over the past two decades has revealed a few key strategies for
visual odometry (VO) and visual simultaneous localization and mapping (VSLAM).
Efforts have focused on the accuracy and robustness of pose tracking
\cite{MonoSLAM,PTAM,ORBSLAM,engel2014lsd,SVO2017,DSO2017} 
and mapping \cite{newcombe2011kinectfusion, whelan2016elasticfusion}, 
while meeting the real-time requirement (e.g. 30 fps) on desktops \& laptops.  
However, the compute resources on practical robotics and AR platforms is more
diverse, and somtimes more limiting.  
When targeting diverse platforms, VO/VSLAM should be accurate and robust while
exhibiting low-latency, i.e., the time cost from capturing an image to
estimating the corresponding pose should be low.   

\begin{figure*}[!htb]
  \centering
  \includegraphics[width=0.4\linewidth]{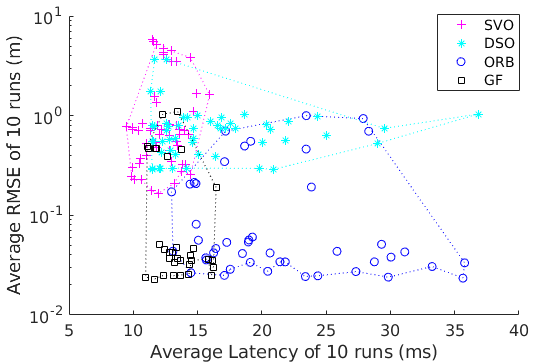}
  \hspace{5pt}
  \includegraphics[width=0.45\linewidth]{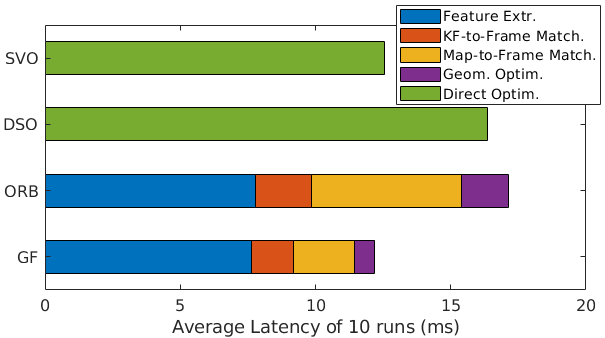}
  \caption{Latency reduction and accuracy preservation of proposed approach on
  EuRoC MAV benchmark.  Four monocular VO/VSLAM systems are assessed:
  semidirect SVO\cite{SVO2017}, direct DSO\cite{DSO2017}, feature-based
  ORB\cite{ORBSLAM}, and proposed GF-ORB.  
  \textbf{Left}: Latency vs. accuracy of 4 systems.  The working region 
  of each system (dashed contour) is obtained by adjusting the maximum number
  of features/patches per frame. 
  \textbf{Right}: Latency break down averages per module in pose tracking
  pipelines on the EuRoC benchmark.  A configuration that yields good
  trade-off of latency and accuracy is set: 
  800 features/patches extracted per frame; 
  for GF-ORB the good features threshold is 100.  
  \label{fig:Illustration}} 	
   \vspace*{-0.5cm}
\end{figure*}

Dedicated hardware improves the runtime of VO/VSLAM on compute-constrained 
platforms.  FPGA-based image processing speeds up feature extraction 
\cite{fang2017fpga, quigley2018open}, which is a dominant computation for
feature-based methods (see Fig. \ref{fig:Illustration}, right).
Exploring the co-design space between VO (with inertial) algorithm and
hardware illuminates parametric settings that improve VO 
output \cite{zhang2017visual}.  
Building more efficient VO/VSLAM algorithms, in parallel with better hardware
integration, is important to realizing the goal of accurate, low-latency VSLAM.  
{\em The focus of this paper is on algorithm design aspects of modern VSLAM.} 
As an alternative sensing approach, low-latency visual sensors such as event
camera have also been studied for VO/VSLAM tasks 
\cite{kueng2016low, rebecq2017real, zhu2017event}.  
Application contexts, however, may require that more traditional visual
cameras be used.
Frame-based cameras are widely recognized as the primary vision sensor
in a generic VO/VSLAM system (and downstream detection/recognition
systems).  The majority of VO/VSLAM systems are designed for frame-based
cameras.

State-of-the-art VO/VSLAM systems on frame-based cameras break down into
three groups: feature-based, direct, and semidirect systems.  
Feature-based VO/VSLAM typically consists of two modules: 
data association (i.e. feature extraction \& matching) and state optimization.  
Due to robust and repeatable modern point-feature descriptors 
\cite{calonder2010brief, leutenegger2011brisk, rublee2011orb, vandergheynst2012freak},
feature-based systems 
(e.g., ORB-SLAM \cite{ORBSLAM}, OKVIS \cite{leutenegger2015keyframe}) 
benefit from long-baseline feature matchings, and are accurate and robust 
in most scenarios with sufficient visual textures.
In low-texture scenarios where point features fail, line features may be
reliable alternative features for VO/VSLAM 
\cite{pvribyl2016camera,PLSLAM,GoodLine}. 
However, feature-based VO/VSLAM typically has high latency: data association
is computationally expensive.
Direct VO/VSLAM systems such as \cite{engel2014lsd, DSO2017}, omit the
explicit data association module and optimize a direct objective defined on
the image measurements.  
In general, the computational load and latency of direct systems are lower
than feature-based systems.  
However, the underlying direct objective is non-convex \& non-smooth,
and therefore harder to optimize versus the geometric objective used in
feature-based systems.  
Furthermore, immediate recovery from track failure (i.e. relocalization) 
is a known issue for direct systems.  
Therefore, direct systems require certain conditions 
\cite{DSO2017,yang2018challenges,schubert2018direct} for optimal
performance, e.g.,
global shutter camera with precise calibration, 
minor or slow changes in lighting conditions, 
accurate motion prediction or smooth \& slow camera motion. 
These conditions limit the applicability of direct systems to many robotics
and AR applications, where VO/VSLAM is expected to operate with noisy sensory
input under changing environments, for long duration.  
In addition, direct measurements rarely persist over long-baselines.
For applications with frequent revisits, the percentage of long-baseline
associations utilized by direct systems is lower than feature-based ones,
impacting the performance of direct VO/VSLAM.
Semidirect systems \cite{SVO2017} also leverage direct measurements for pose
tracking, thereby inheriting the reduced tracking performance property
relative to feature-based methods.  
To summarize, there is a gap in the middle ground between performance 
(accuracy \& robustness) and efficiency (low-latency) for state-of-the-art
VO/VSLAM systems: 
feature-based systems have good performance yet the latency can be quite
high due to the explicit data association; direct \& semidirect systems 
have less latency than feature-based ones, however they are either 
inapplicable in many target scenarios, or exhibit relatively lower accuracy or
robustness.

The objective of this research to balance latency and performance with a
computational enhancement applied to feature-based VO/VSLAM. The enhancement
reduces the latency to the level of direct systems while preserving the
accuracy and robustness of feature-based ones.  The key observation that
renders the objective achievable being: not all the feature matchings
contribute equally to the accurate \& robust estimation of camera pose.  
If there is an efficient approach to identify a small subset of features that
are most valuable towards pose estimation, a.k.a. {\em good features}, then
both data association and state optimization should utilize the good features
only.  Latency of pose tracking will be significantly reduced while preserving
accuracy \& robustness.

The primary outcome of this work is illustrated in Fig. \ref{fig:Illustration}.  
The left column depicts the latency-accuracy trade-off of 4 monocular VO/VSLAM
systems on a public benchmark (EuRoC MAV \cite{burri2016euroc}) by plotting
the operational domain of the systems.
Each marker on the plot represents a successfully tracked sequence (zero track
loss in 10 repeated trials) for the denoted VO/VSLAM system.
To better understand the latency-accuracy trade-off in each VO/VSLAM system,
we adjust the maximum number of features/patches extracted per frame (for
GF-ORB, we also adjust the maximum number of good feature being matched per
frame), 
to obtain the working region of each system in the latency-accuracy
plot (in dashed contour).  
In Fig. \ref{fig:Illustration}, feature-based ORB-SLAM occupies the lower-right
portion; it is accurate with high-latency.  Direct DSO achieves lower
latencies under some configurations, but it has an order of magnitude higher
{\em absolute root-mean-square error} (RMSE) than ORB-SLAM; 
semidirect SVO has a tighter-bounded working region at the upper-left, meaning
it is efficient yet inaccurate.  
The objectives of low-latency and high-accuracy are achieved with the proposed
approach, GF-ORB-SLAM, whose markers are located in the lower-left region of
the plot.
Fig. \ref{fig:Illustration} presents the break down of latency introduced by
each module in pose tracking pipelines, under a typical configurations for
each VO/VSLAM system.  When GF-ORB-SLAM is compared with the baseline
ORB-SLAM, the time cost of feature extraction is identical, but the feature
matching and subsequent modules' costs are significantly reduced.
The latency of GF-ORB is the lowest among all four systems. 

This paper extends our previous work on good features \cite{zhao2018good2}.  
Extensions include an in-depth study of randomized acceleration (Sec.
\ref{sec::lazier}), and the addition of feature selection with active matching
(Sec.  \ref{sec::goodmatch}).  Further, the experiments (Sec. \ref{sec::exp})
are more comprehensive.  Contributions of this work include:
\smallskip

\noindent
1) Study of the \textbf{error model} of least squares pose optimization, 
  to connect the performance of pose optimization to the spectral
  property of a weighted Jacobian matrix; \\
2) \textbf{Exploration of metrics} connected to the least squares conditioning of
  pose optimization, with quantification of {\em Max-logDet} as the optimal
  metric; \\ 
3) Introduction of an \textbf{efficient good feature selection algorithm}
  using the {\em Max-logDet} metric, which is an order of magnitude faster
  than state-of-the-art feature selection approaches; \\
4) Fusion of good feature selection and active matching into a
  \textbf{generic good feature matching algorithm}, which is efficient and
  applicable to feature-based VO/VSLAM; and  \\
5) \textbf{Comprehensive evaluation} of the proposed good feature matching on 
  a state-of-the-art feature-based VSLAM system, with multiple benchmarks, 
  sensor setups, and compute platforms.  Evaluation results
  demonstrate both latency reduction and accuracy \& robustness preservation
  with the proposed method.  
  Both 
  monocular\footnote{\url{https://github.com/ivalab/GF_ORB_SLAM}} 
  and stereo\footnote{\url{https://github.com/ivalab/gf_orb_slam2}}
  SLAM implementations are open-sourced.

\section{Related Work} \label{sec::liter}
This work is closely connected to following three research topics in 
robotics and computer vision: feature selection, submatrix selection, and
active matching.  In what follows, we review the literatures 
in each topic, and discuss the connection between existing works and 
this paper.

\subsection{Feature Selection}
Feature selection has been widely applied in VO/VSLAM for performance 
and efficiency purposes.  Conventionally, fully data-driven methods 
such as random sample consensus (RANSAC) \cite{MonoSLAM} and 
joint compatibility branch and bound (JCBB) \cite{JCBB} are used 
to reject outlier features.  The computational efficiency of 
these methods are improved in extended works \cite{KalmanSAC,1pRANSAC}.  
These outlier rejection methods are utilized in
many VO/VSLAM systems \cite{MonoSLAM,PTAM,ORBSLAM} to improve 
the robustness of state estimation.  

Apart from outlier rejection, feature selection methods are 
also been utilized for inlier selection, which aims to 
identify valuable inlier matches from useless ones.  One major 
benefit of inlier selection is the reduction of computation 
(and latency thereafter), since only a small set of selected inliers 
are processed by VO/VSLAM.  In addition, it is possible to 
improve accuracy with inlier selection, as demonstrated in 
\cite{cvivsic2015stereo, zhang2015optimally, zhang2015good, zhao2018good2}.  
The scope of this paper is on inlier selection, which reduces the 
latency of VO/VSLAM while preserving the accuracy and robustness.

Image appearance has been commonly used to guide inlier selection:
feature points with distinct color/texture patterns are more likely to get
matched correctly \cite{GoodFeaturesToTrack,sala2006landmark,shi2013feature}.  
However, these works solely rely on quantifying distinct appearance,
while the structural information of the 3D world and the camera motion
are ignored.  Appearance cues are important in feature selection, however, the
focus of this paper is on the latter properties:
identifying valuable features based on structural and motion information.  
The proposed structure-driven method can combine with 
a complementary appearance-based approach.

To exploit the structural and motion information, covariance-based inlier 
selection methods are studied 
\cite{MonoSLAM,covarianceRecovery,zhang2005entropy,lerner2007landmark,cheein2009feature,carlone2019attention}.  
Most of these works are based on pose covariance matrix, which has two key characteristics: 
1) it contains both structural and motion information implicitly, and 
2) it approximately represents the uncertainty ellipsoid of pose estimation.  
Based on the pose covariance matrix, different metrics were introduced
to guide the inlier selection, such as information gain \cite{MonoSLAM}, 
entropy \cite{zhang2005entropy}, trace \cite{lerner2007landmark}, 
covariance ratio \cite{cheein2009feature}, minimum eigenvalue and log determinant \cite{carlone2019attention}.    
Covariance-based inlier selection methods are studied for 
both filtering-based VO/VSLAM 
\cite{MonoSLAM,covarianceRecovery,zhang2005entropy,lerner2007landmark,cheein2009feature} 
and BA-based VO/VSLAM \cite{ila2017fast,ila2017slam++,carlone2019attention}.  

The observability matrix has been studied as an alternative to the covariance
matrix for guiding feature selection \cite{zhang2015optimally, zhang2015good}.  
In these works, the connection 
between pose tracking accuracy and observability conditioning of SLAM 
as a dynamic system is studied.  The insight of their work being: the better
conditioned the SLAM system is, the more tolerant the pose estimator
will be towards feature measurement error.  To that end, the minimum singular
value of the observability matrix is used as a metric to guide feature 
selection.  However, observability matrix can only be constructed efficiently 
under piecewise linear assumption, which limits the applicability of 
observability-based feature selection.  Furthermore, we argue that 
covariance matrix is better suited for the static or {\em instantaneous} bundle
adjustment (BA) problem as formulated in pose tracking, as it can be
constructed efficiently for non-linear optimizers under a sparsity assumption.

The study in \cite{carlone2019attention} is most related to our work.  
In \cite{carlone2019attention}, feature selection is performed by
maximizing the information gain of pose estimation within a prediction horizon.
Two feature selection metrics were evaluated, minimal eigenvalue and log
determinant ({\em Max-logDet}). 
Though the log determinant metric is utilized in our work, the 
algorithm for approximately selecting the feature subset maximizing {\em
logDet} differs, as well as the matrix whose conditioning is optimized.  
Compared with \cite{carlone2019attention}, our work is more applicable 
for low-latency pose tracking from two key advantages.  
First, the lazier-greedy algorithm presented in our paper is efficient.
It takes an order of magnitude less time than the lazy-greedy algorithm 
of \cite{carlone2019attention}, yet preserves the optimality bound.  
Second, we present the combination of efficient feature selection 
and active feature matching, which reduces the latency of both data 
association and state optimization.  Meanwhile, \cite{carlone2019attention} 
selects features after data association, therefore leaving the latency 
of data association unchanged.  The experimental results in \cite{carlone2019attention}
support these claims: there are occasions when feature selection actually
increases the latency of full pipeline, compared with the original all-feature
approach.

\subsection{Submatrix Selection}
A key perspective of this work is connecting feature selection 
with submatrix selection under a spectral preservation objective, 
which has been extensively studied in the fields of computational 
theory and machine learning  
\cite{gu1996efficient,boutsidis2009improved,shamaiah2010greedy,jawaid2015submodularity,summers2016submodularity}.
Submatrix selection is an NP-hard, combinatorial optimization problem. 
To make submatrix selection more amendable to optimization, one structural
property, {\em submodularity}, has been explored 
\cite{shamaiah2010greedy,jawaid2015submodularity,summers2016submodularity}.  
If a set function (e.g. matrix-revealing metric in this paper) is 
submodular and monotone increasing, then the combinatorial optimization 
of the set function (e.g. submatrix selection in this paper) can be 
approximated by simple greedy method with approximation guarantee.  

Compared with deterministic methods (e.g. classic greedy), randomized submatrix 
selection has been proven to be a faster alternative with probabilistic performance 
guarantees \cite{drineas2008relative,drineas2012fast}.  Combining randomized selection 
with a deterministic method yields fast yet near-optimal submatrix selection,
as demonstrated for specific matrix norms 
\cite{boutsidis2009improved,boutsidis2014near} 
and general submodular functions 
\cite{mirzasoleiman2015lazier,hassidim2017robust}.
This paper uses the ideas from these works to design a good feature selection
algorithm.

\subsection{Active Matching}
Another key perspective of this work is combining feature 
selection algorithm with active feature matching, which leads 
to latency-reduction in both data association and state optimization.  
Active matching refers to the guided feature matching methods 
that prioritize processing resource (e.g. CPU percentage, latency budget) on 
a subset of features.  Compared with brute force approaches that treat all
features equally, active matching is potentially more efficient, especially
under resource constraints.  

Active matching has been primarily studied for filter-based VO/VSLAM, with 
representative works \cite{activeSearch,activeMatching,scalableActiveMatching}. 
Traditional active matching methods require dense covariance matrices 
(i.e. majority of off-diagonal components are filled), 
and are less relevant to modern VO/VSLAM systems driven by non-linear sparse
optimizers.  Furthermore, the algorithms used by these active matching methods
were computate-heavy, and provided little benefit when integrated into
the real-time pose tracking thread of modern VO/VSLAM system.  
Therefore, the idea of active matching became less attractive. Quoting 
\cite{davison2018futuremapping}: 
``the problem with this idea (active searching) was that ... too much
computation is required to decide {\em where} to look.''
In this paper, we demonstrate the worth of revisiting the classic idea of
active matching: 
the proposed good feature matching algorithm is extremely efficient and
applicable, based upon specific matrices and selection algorithm tailored for
non-linear optimization.  
To the best of our knowledge, this is the first work
to demonstrate the applicability of latency-reduction and accuracy preservation
in {\em real-time pose tracking} with active feature selection.
The benefit of active matching is realized because the
structure of modern VO/VSLAM methods permits first asking whether it is
desirable to actively look, then to determine where.  In effect, it decides {\em
when} to look, {\em how much} to look, and {\em where} to look.

\section{Least Squares Pose Optimization Uncertainty} \label{sec::prelim}
This section examines pose covariance as a function of measurement
and point estimation error, with reference to the least squares pose 
optimization objective commonly used in feature-based VO/VSLAM.  
The intent is to identify what matrices influence the pose covariance.
Without loss of generality, write the least squares objective as,
\begin{equation}  \label{eq:LeastSquare}
  \min \left\Vert h(x, p)-z \right\Vert^2,
\end{equation}
where $x$ is the pose of the camera, 
$p$ are the 3D feature points and 
$z$ are the corresponding 2D image measurements.  
The measurement function, $h(x, p)$, is a combination of the $SE(3)$
transformation (world-to-camera) and pin-hole projection.  
For simplification, we omit camera lens distortion in $h(x, p)$.
Correcting for lens distortion involves undistorting the image measurements,
$z$, based on the camera calibration parameters so that the model given by $h(x,p)$
is valid.  We base the theory of good feature selection upon the objective of
\eqref{eq:LeastSquare}.

Solving the least squares objective often involves the first-order
approximation to the non-linear measurement function $h(x, p)$:
\begin{equation} \label{eq:FirstOrderApprox}
  \left\Vert h(x, p) -z \right\Vert^2 
    {\approx} \left\Vert h(x^{(s)}, p) + H_x(x-x^{(s)}) - z \right\Vert^2,
\end{equation}
where $H_x$ is the measurement Jacobian linearized about the initial guess 
$x^{(s)}$.  
To minimize of the first-order approximation Eq~\eqref{eq:FirstOrderApprox}
via Gauss-Newton, the pose estimate is iteratively updated via
\begin{equation} \label{eq:GN_oneIter}
  x^{(s+1)} = x^{(s)} {+} H_x^+( z-h(x^{(s)}, p) ),
\end{equation}
where $H_x^+$ is the left pseudoinverse of $H_x$.

The accuracy of Gauss-Newton depends on the residual error $\epsilon_r$, which 
can be decomposed into two terms: measurement error $\epsilon_z$ and map error 
$\epsilon_p$.  Using the first-order approximation of $h(x, p)$ at the 
estimated pose $x^{(s)}$ and map point $p$ to connect the pose 
optimization error with measurement and map errors leads to
\begin{equation} \label{eq:Error_Formula}
  \epsilon_x = H_x^+ \epsilon_r = H_x^+( \epsilon_z - H_p \epsilon_p ).
\end{equation}
The Jacobian of map-to-image projection, $H_p$, is a diagonal matrix
with $n$ diagonal blocks ${H_p(i)}$, where $n$ is the number of matched
features.  

The Bundle Adjustment literature commonly assumes that the measurement
error follows an independent and identically distributed Gaussian 
(i.e., there is an i.i.d. assumption).  While keeping the independent 
Gaussian assumption, this paper relaxes the identical assumption.  
Instead, the distribution of measurement error is correlated with the
image processing parameters, e.g. the scale-level of the extracted keypoint.  
Without loss of generality, the measurement error and the map error are
modeled as 
$\epsilon_z(i) \sim N(0, \Sigma_z(i))$ and 
$\epsilon_p(i) \sim N(0, \Sigma_p(i))$.
The combined residual error on image plane follows
$\epsilon_r(i) \sim N(0, \Sigma_r(i))$, where 
\begin{equation} \label{eq:Residual_Formula}
  \Sigma_r(i) = \Sigma_z(i) + H_p(i) \Sigma_p(i) H_p(i)^T.
\end{equation}
Applying a Cholesky decomposition to each $2\times 2$ covariance matrix 
$\Sigma_r(i)$ leads to $\Sigma_r(i) = W_r(i) W_r(i)^T$.  
Assembling $W_r(i)$ from all $n$ residual terms into a $2n\times
2n$ block diagonal weight matrix $W_r$ and linking to the pose covariance 
matrix,
\begin{equation} \label{eq:Pose_Cov}
  \Sigma_x = H_x^+ \Sigma_r (H_x^+)^{T} = H_x^+ W_r (H_x^+ W_r)^{T}.
\end{equation}
We aim to simplify the right hand side.  Moving both matrices on the right hand
side of Eq~\eqref{eq:Pose_Cov} to the left hand side,
\begin{equation}
	W_r^{-1} H_x \Sigma_x (W_r^{-1} H_x)^T = {\bf I}.
\end{equation}
Note that $W_r^{-1}$ is still a block diagonal matrix, consisting of 
$2\times 2$ blocks denoted by $W_r^{-1}(i)$.  Meanwhile, each row block 
of measurement Jacobian $H_x$ can be written as $H_x(i)$. 
Following through on the block-wise multiplication results in the matrix
$H_c$:
\begin{equation} \label{eq:Matrix_Merge}
	H_c = \left[
		\begin{array}{c}
		W_r^{-1}(0) H_x(0) \\
		... \\
		W_r^{-1}(n-1) H_x(n-1)
		\end{array}
	\right], 
\end{equation}
from which the simplified pose covariance matrix follows:
\begin{equation} \label{eq:Pose_Cov_Final}
  \Sigma_x = H_c^+ (H_c^+)^{T} = (H_c^{T} H_c)^{-1},
\end{equation}
assuming that there are sufficient tracked map points so that $H_c$ is
full rank.
The conditioning of $H_c$ determines the error propagation properties of
the iteratively solved least-squares solution for the camera pose $x$.

\section{Good Feature Selection Using Max-LogDet} \label{sec::goodfeat}
The pose covariance matrix $\Sigma_x$ represents the uncertainty ellipsoid in 
pose configuration space.  According to Eq~\eqref{eq:Pose_Cov_Final}, one 
should use all the features/measurements available to minimize the
uncertainty (i.e. variance) of pose estimation: with more measurements,
the singular values of $H_c$ should increase in magnitude.  
The worst case uncertainty would be proportional to the inverse of 
minimal singular value $\sigma_{min}(H_c)$, whereas in the best case it would be 
proportional to the inverse of maximal singular value $\sigma_{max}(H_c)$. 

However, for the purpose of low-latency pose tracking, one should only
utilize {\bf sufficient} features.  There is a tension between latency
and error rejection.
From the analysis, the uncertainty of least squares pose optimization
problem is bounded by the extremal spectral properties of the matrix $H_c$.
Hence, one possible metric to measure the sufficiency of a feature
subset would be the factor of the worst case scenario $\sigma_{min}(H_c)$.  
Meanwhile, one may argue that the extremal spectral properties only decides the 
upper and lower bounds of pose optimization uncertainty.  
The true values would depend on what the overall spectral properties of the system are. 
It follows then, that another possible measurement of sufficiency would
be the overall spectral properties of $H_c$.  

Define the {\em good feature selection} problem to be: 
Given a set of 2D-3D feature matchings, find a constant-cardinality subset 
from them, such that the error of least squares pose optimization is 
minimized when using the subset only.
Based on the previous discussion, the good feature
selection problem is equivalent to submatrix selection: 
Given a matrix $H_c$, select a subset of row blocks so that the overall 
spectral properties of the selected submatrix are preserved as much as 
possible.

\paragraph*{A Note Regarding Good Feature Selection \& Matching}
Good feature selection is slightly different from the final goal of 
this work, good feature matching.  In good feature selection, all 
2D-3D feature matchings are assumed to be available in the first place.  
In good feature matching, only the 3D features are known 
in the beginning, while the 2D-3D matchings are partially 
revealed during the guided matching process.  Still, these two problems 
share the same core, which is how to prioritize a subset of features 
over the others for accuracy-preserving purposes.  
The section following this one will describe how to translate a good feature
selection solution to a good feature matching solution. 

\subsection{Submodularity in Submatrix Selection}
Submatrix selection with spectral preservation has been extensively
studied in the numerical methods and machine learning fields 
\cite{gu1996efficient,boutsidis2009improved}, for
which several matrix-revealing metrics exist to score the subset
selection process.  They are listed in Table~\ref{tab:Reveal-Metric}.
Subset selection with any of the listed matrix-revealing metrics is
equivalent to a finite combinatorial optimization problem 
with a cardinality constraint:
\begin{equation} \label{eq:Combine_Opt}
  \max_{S\subseteq\{1,2,...,n\}, |S|=k} f([H_c(S)]^T [H_c(S)])
\end{equation}
where $S$ contains the index subsets of selected row blocks from the full 
matrix $H_c$, $[H_c(S)]$ is the corresponding row-wise concatenated submatrix, 
$k$ is the cardinality of subset, and $f$ the matrix-revealing metric.

While the combinatorial optimization can be solved by brute force, 
the exponentially-growing problem space quickly becomes impractical 
to search over for real-time VO/VSLAM applications.  
To employ more efficient subset selection strategies while limiting the
loss in optimality, one structural property of the problem may be
exploited, submodularity
\cite{summers2016submodularity,jawaid2015submodularity,shamaiah2010greedy,carlone2019attention}.  
If a set function (e.g. matrix-revealing metric) is
submodular and monotone increasing, then the combinatorial optimization
of the set function (e.g. subset selection) found via greedy methods has 
known approximation guarantees.

Except for {\em Min-Cond}, the metrics listed in Table~\ref{tab:Reveal-Metric} 
are either submodular or approximately submodular, and monotone increasing.
The {\em Max-logDet} metric is submodular \cite{shamaiah2010greedy},
while the {\em Max-Trace} is modular (a stronger property) 
\cite{summers2016submodularity}.  
Lastly, {\em Max-MinEigenValue} is approximately submodular 
\cite{jawaid2015submodularity}.
Therefore, selecting row blocks (as well as the corresponding features)
with these metrics can be approximately solved with greedy methods.
Using these known properties, the aim here is to 
arrive at an efficient algorithm for performing good feature selection
or matching without significant loss in optimality.

\begin{table}[t]
\begin{center}
\begin{tabular}{ l | l }
{\em Max-Trace}				&	Trace $Tr(Q) = \sum_{1}^{m} Q_{ii}$ is max.	\\
{\em Min-Cond}				&	Condition $\kappa(Q) = \lambda_{1}(Q)/\lambda_{m}(Q)$ is min.	\\
{\em Max-MinEigenValue}		&	Min. eigenvalue $\lambda_{m}(Q)$ is max.	\\
{\em Max-logDet}			&	Log. of determinant $\log \det(Q)$ is max.	\\
\end{tabular}
\end{center}
\caption{Commonly used matrix-revealing metrics, with input square matrix $Q$ of rank $m$.} \label{tab:Reveal-Metric}
\end{table}

\begin{figure}[t]
\centering
\includegraphics[width=0.4\linewidth]{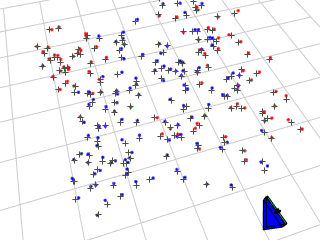} \hspace{15pt}
\includegraphics[width=0.4\linewidth]{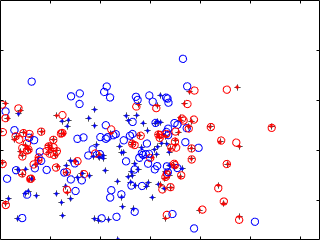}
\caption{Simulated pose optimization scenario.  
\textbf{Left}: map view. \textbf{Right}: camera view.  
Selected features are in red, while unselected ones are in blue. \label{fig:Simu_PnP_Scene}} 	
\vspace{-0.5cm}
\end{figure}

\subsection{Simulation of Good Feature Selection}

To explore which matrix-revealing metrics might best guide good
feature/row block selection for least squares pose optimization, this
section evaluates the candidate metrics via simulation.  
The Matlab simulation environment \cite{LandmarkImpact}, which assumes perfect 
data association, provides the testing framework.  The evaluation scenario is 
illustrated in Fig. \ref{fig:Simu_PnP_Scene}.  The camera/robot is
spawned at the origin of the world frame, and a fixed number  of 3D
feature points are randomly generated in front of the camera (200 in
this simulation).  
After applying a small random pose transform to the robot/camera, the 2D 
projections of feature points are measured and perfectly matched with known 
3D feature points.  Then a Gauss-Newton optimizer uses the matchings to 
estimate the pose transform.

\begin{figure*}[!htb]
	\centering
	\includegraphics[width=0.32\linewidth]{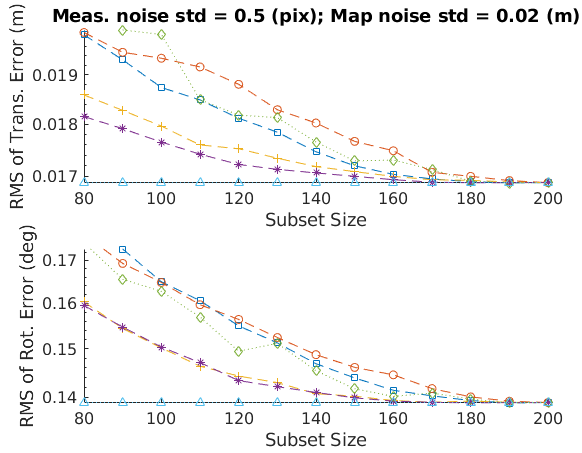}
	\includegraphics[width=0.32\linewidth]{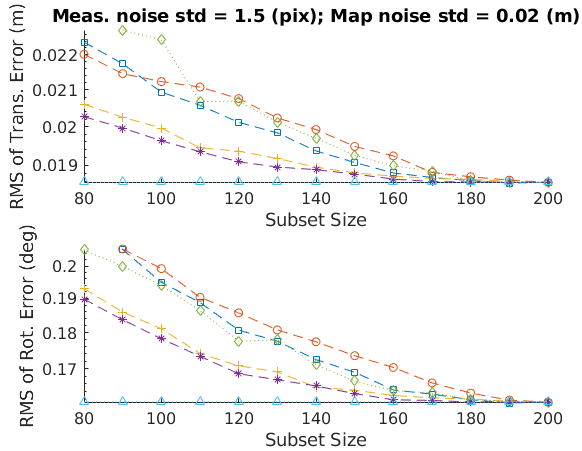}
	\includegraphics[width=0.32\linewidth]{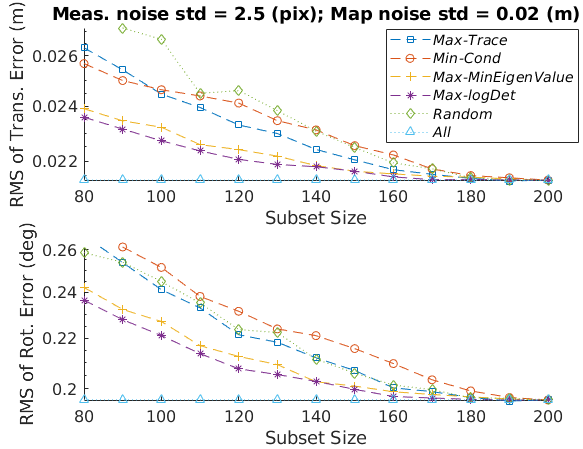}\\
	\caption{Simulation results of least squares pose optimization.  
	First row: RMS of translational error under 3 levels of residual error.
	Second row: RMS of rotational error under 3 levels of residual error.  \label{fig:Simu_PnP_Res}}
	\vspace{-0.5cm}
\end{figure*}

To simulate the residual error, both the 3D mapped features and the 2D
measurements are perturbed with noise.  A zero-mean Gaussian with the 
standard deviation of 0.02m are added to the 3D features stored as map.
Three levels of measurement error are added to 2D measurements:
zero-mean Gaussian with standard deviation of 0.5, 1.5 and 2.5 pixel.  
Subset sizes ranging from 80 to 200 are tested.  To be statistically
sound, 300 runs are repeated for each configuration.

A feature selection module is inserted prior to Gauss-Newton pose 
optimization, so that only a subset of selected features is sent into 
the optimizer.  
Feature selection is done in two steps: 
1) compute the combined matrix $H_c$ 
from measurement Jacobian $H_x$ and noise weighting matrix $W_r$, 
2) greedy selection of row block $H_c(i)$ based on the matrix-revealing 
metric, until reaching the chosen subset size.  
The simulation results are presented in Fig. \ref{fig:Simu_PnP_Res}, 
with the root-mean-square (RMS) of translational error (m) in the first row 
and rotational error (deg) in the second row.  
Each of the matrix-revealing metrics in Table~\ref{tab:Reveal-Metric} is tested.
For reference, the plots include simulation results with randomized
subset selection ({\em Random}) and with all features available 
({\em All}).

From Fig. \ref{fig:Simu_PnP_Res}, two metrics stand out: {\em Max-MinEigenValue} and 
{\em Max-logDet}.  Under all residual noise levels, their curves more
quickly approach the baseline error ({\em All}) as a function of the
subset size.  
Based on the outcomes, {\em Max-logDet} is chosen as the metric to guide good 
feature selection.  The reasons being:  
(1) 
According to Fig. \ref{fig:Simu_PnP_Res}, the error curves of {\em
Max-logDet} are always lower, if not at the same level, than those of
{\em Max-MinEigenValue}. 
The subset selected with {\em Max-logDet} approximates the original 
full feature set better than the subset with {\em Max-MinEigenValue}. 
As discussed previously, greedy selection with {\em Max-logDet} has 
guaranteed approximation ratio due to submodularity.
(2) The computational cost of {\em logDet} is lower than that
of {\em MinEigenValue}.  
The main {\em logDet} computation is Cholesky factorization, with
a complexity of $\mathcal{O}(0.33 n^3)$, whereas for  
{\em MinEigenValue} the complexity is 
$\mathcal{O}(22 n^3)$ \cite{golub2012matrix}. 
Lastly, the error rate of {\em Random} selection is 
much higher than {\em logDet}-guided selection. To be effective randomized 
selection requires a larger subset size.

\section{Efficient Good Feature Selection} \label{sec::lazier}
Subset selection with {\em Max-logDet} metric has been studied for 
sensor selection \cite{shamaiah2010greedy} and feature selection 
\cite{carlone2019attention}, with reliance on a simple greedy algorithm 
commonly used to approximate the original NP-hard combinational optimization
problem.  Since {\em Max-logDet} is submodular and monotone increasing, the
approximation ratio of a greedy approach is proven to be $1-1/e$ 
\cite{summers2016submodularity}.  This approximation ratio is the best
achievable by any polynomial time algorithm under the assumption that 
$P \neq NP$.

The classical greedy algorithm can be enhanced into an accelerated 
version, lazy greedy \cite{minoux1978accelerated}.  
Instead of evaluating the actual margin gain of the chosen metric (e.g. {\em
logDet}) at each iteration, a computationally-cheap, approximate upper bound
is estimated and utilized to reject unwanted blocks/features. 
Speed enhancement of the lazy greedy hinges on the tightness of the upper
bound.  
Consider an idealized case, where computing the upper bound 
takes zero-cost and a constant rejection ratio $\rho$ is achieved with 
the upper bound.  Hence the total complexity of selecting $k$ features
given $n$ candidates using lazy greedy algorithm is $\mathcal{O}(k(1-\rho)n)$: 
the lazy greedy algorithm has to run $k$ rounds, in each 
round it will go through $(1-\rho)n$ candidates to identify the 
current best feature.  

Unfortunately, the commonly used upper bound of {\em logDet}, as derived 
from Hadamard's inequality \cite{horn1990matrix}, is quite loose 
\cite{carlone2019attention} (i.e. $\rho \approx 0$):
\begin{equation} \label{eq:LogDet_Upper}
  \log \det(Q)\leq \sum_{i=1}^{m} \log(Q_{ii}),\ \rank(Q)=m. 
\end{equation}
Therefore, {\em Max-logDet} feature selection does not appreciably benefit 
from a lazy greedy implementation.  
As reported in \cite{carlone2019attention} and further confirmed in the
simulation to be discussed shortly, the time cost of lazy greedy selection
exceeds the real-time requirement (e.g. 30ms per frame), therefore
lazy-greedy with {\em logDet} and {\em Trace} is impractical for good
feature selection in real-time VO/VSLAM applications.

\begin{figure*}[!htb]
	\centering
	\includegraphics[width=\linewidth]{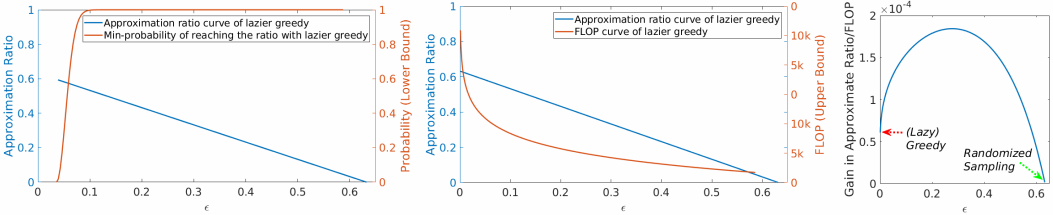}
	\caption{Illustration of performance \& efficiency of lazier greedy, when selecting 450 subset from 1500 rows with 
	average approximation ratio $\mu = 0.8$ in maximizing margin gain ({\em logDet}).  
	\textbf{Left}: Approximation ratio \& probabilistic guarantee of lazier greedy, w.r.t. different decay factor $\epsilon$.
	\textbf{Middle}: Approximation ratio \& computation cost (FLOP) of lazier greedy, w.r.t. different decay factor $\epsilon$.
	\textbf{Right}: Efficiency of lazier greedy, w.r.t. different decay factor $\epsilon$.  \label{fig:Theo_Lazier_Effi}}
	\vspace{-0.5cm}
\end{figure*}

\subsection{Lazier-than-lazy Greedy}
To speed up the greedy feature selection, we explore the combination
of deterministic selection (e.g. lazy greedy algorithm) and randomized 
acceleration (e.g. random sampling).  
One well-recognized method of 
combining these two, is lazier-than-lazy greedy \cite{mirzasoleiman2015lazier} 
(referred as lazier greedy in the following).  
The idea of lazier greedy is simple: 
at each round of greedy selection, instead 
of going through all $n$ candidates, only a random subset of candidates are 
evaluated to identify the current best feature.  Furthermore, the size of 
the random subset $s$ can be controlled by a decay factor $\epsilon$: 
$s=\frac{n}{k}\log(\frac{1}{\epsilon})$.  In this way, the total complexity is reduced from
$\mathcal{O}(k(1-\rho)n)$ to $\mathcal{O}(\log(\frac{1}{\epsilon})n)$.  
Importantly, lazier greedy is near-optimal:
\begin{theorem} \cite{mirzasoleiman2015lazier}
\label{optimal_in_exp}
Let $f$ be a non-negative monotone submoduar function. 
Let us also set $s=\frac{n}{k}\log(\frac{1}{\epsilon})$. 
Then lazier greedy achieves a $\left( 1-1/e-\epsilon \right)$ approximation
guarantee in expectation to the optimum solution of problem in
Eq~\ref{eq:Combine_Opt}.
\end{theorem} 
\begin{theorem} \cite{hassidim2017robust}
\label{optimal_in_prob}
The expectation of approximation guarantee of $\left( 1-1/e-\epsilon \right)$
is reached with a minimum probability of $1-e(-0.5k(\sqrt{\mu}+ln(\epsilon+e^{-1})/\sqrt{\mu})^2)$, 
when maximizing a monotone submodular function under cardinality constraint $k$
with lazier-greedy.  $\mu \in (0,1]$ is the average approximation ratio 
in maximizing margin gain at each iteration of lazier greedy.
\end{theorem}

The symbols \& formulations in Theorem~\ref{optimal_in_prob} are adjusted from
the original ones \cite{hassidim2017robust} to be consistent with
Theorem~\ref{optimal_in_exp}.  According to these two theorems:
1) lazier greedy introduce a linear loss $\epsilon$ to the approximation 
ratio {\em in expectation};
and 2) the expectation of linear-loss approximation ratio can be guaranteed 
with high probability, as illustrated in Fig. \ref{fig:Theo_Lazier_Effi} (left).  
Compared to the theoretical upper bound of approximation ratio, $1-1/e$, 
which no polynomial time algorithm can exceed \cite{summers2016submodularity}, 
lazier greedy only loses a small chunk from it (in expectation \& in probability).   

The approximation ratio and computational speed up of lazier greedy
hinge on the decay factor $\epsilon$. When the decay factor
$\epsilon=0$, the lazier greedy algorithm is the greedy algorithm.
Meanwhile, when the decay factor is $e^{-\frac{k}{n}}$, lazier greedy
becomes randomized sampling (i.e. $s=1$), which has an approximation ratio
of $1-1/e-e^{-\frac{k}{n}}$ in expectation.  
As illustrated in Fig. \ref{fig:Theo_Lazier_Effi} (middle), the approximation
ratio decays linearly with $\epsilon$, while the computational cost (FLOP)
decays logarithmically.  As $\epsilon$ increases the resulting
computational gain outpaces the loss in optimality, until hitting an
inflection point after which the benefit reduces.  By setting $\epsilon$
to a small positive value, e.g. 0.1-0.5 as indicated in
Fig. \ref{fig:Theo_Lazier_Effi} (right), lazier greedy will have a
slightly degraded optimal bound but with a 3-4x higher efficiency than
lazy greedy.  
Alg~\ref{alg:efficient_maxLogDet} describes an efficient algorithm for
good feature selection based on the near-optimal lazier-greedy.

\begin{algorithm}[!htb]
 \KwData{ $H_c=\{H_c(1),\ H_c(2),\ ...\ ,\ H_c(n)\}$, $k$}
 \KwResult{ $H_c^{sub} \subseteq H_c,\ |H_c^{sub}|=k$}

 $H_c^{sub}\gets \emptyset$\;
 \While{$|H_c^{sub}|<k$}{
 	$H_c^R \gets \text{a random subset obtained by sampling }$ 
 	$s=\frac{n}{k}\log(\frac{1}{\epsilon})\text{ random elements from } H_c$\;
 	$H_c(i) \gets \arg\max_{H_c(i) \in H_c^R} \log \det (H_c(i)^T H_c(i)$
 	$+ [H_c^{sub}]^T [H_c^{sub}])$\;
	$H_c^{sub} \gets H_c^{sub}\cup H_c(i)$\;
 	$H_c \gets H_c \setminus H_c(i)$\;
 }
 \Return $H_c^{sub}$.
 \caption{\small Lazier-greedy good feature selection algorithm. \label{alg:efficient_maxLogDet}}
\end{algorithm}

\subsection{Simulation of Lazier Greedy Feature Selection}

To validate the benefits of lazier greedy, and to identify the proper value
of decay factor $\epsilon$, a simulation of good feature selection is
conducted.  A testing process similar to the Matlab one from previous pose
optimization simulation was implemented C++ for speed assessment.  
The two feature selection algorithms tested are: 
lazy greedy \cite{carlone2019attention} 
and lazier greedy (Alg~\ref{alg:efficient_maxLogDet}).  
Like the simulation of pose optimization, a set of randomly-spawned 3D
feature points, as well as the corresponding 2D measurements, are provided
as input.  Gaussian noise is added to both the 3D mapped features and the 2D
measurements.  The perturbed inputs are fed into a matrix building module,
which estimates the combined matrix $H_c$ for submatrix/feature selection.  

\begin{figure*}[!htb]
	\centering
	\includegraphics[scale=0.35]{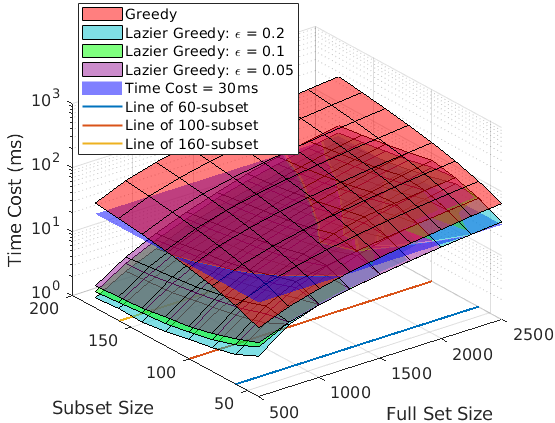}
	\hspace{15pt}
	\includegraphics[scale=0.35]{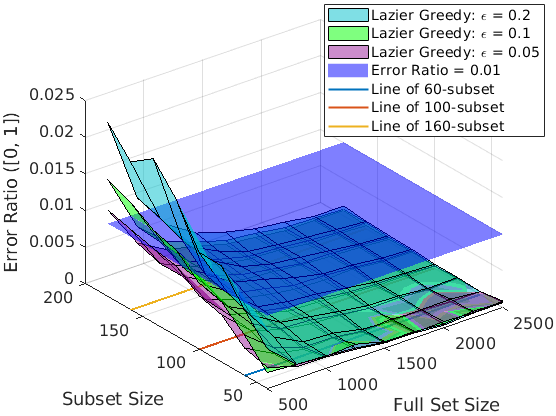}
	\caption{Lazy greedy vs. lazier greedy in feature selection simulation.  
	\textbf{Left}: average time cost of lazy greedy vs. lazier greedy under 
	different decay factor $\epsilon$.
	\textbf{Right}: average error ratio of lazier greedy (compared with lazy 
	greedy baseline; the smaller the better) under different $\epsilon$.  
	Three exemplar working spaces of typical VSLAM problems are plotted (as lines) 
	in both figures, with 60, 100 \& 160 feature subset selected for pose tracking.
	\label{fig:Simu_Lazier_Surf}}
	\vspace{-0.3cm}
\end{figure*}

To assess the performance and efficiency of good feature selection
comprehensively, we sweep through the three parameters:
the size of 3D feature set from 500 to 2500, 
the size of desired feature subset from 40 to 180,
and the decay factor from 0.9 to 0.005. 
For each parameter combination, we randomly spawn 100 different worlds and 
evaluate each feature selection algorithm on each world.  
Due to the randomness of lazier greedy, we repeat it 20 times under each
configuration.

\begin{figure*}[!htb]
	\centering
	\includegraphics[clip, trim=0cm 14.8cm 0cm 0cm, width=\linewidth]{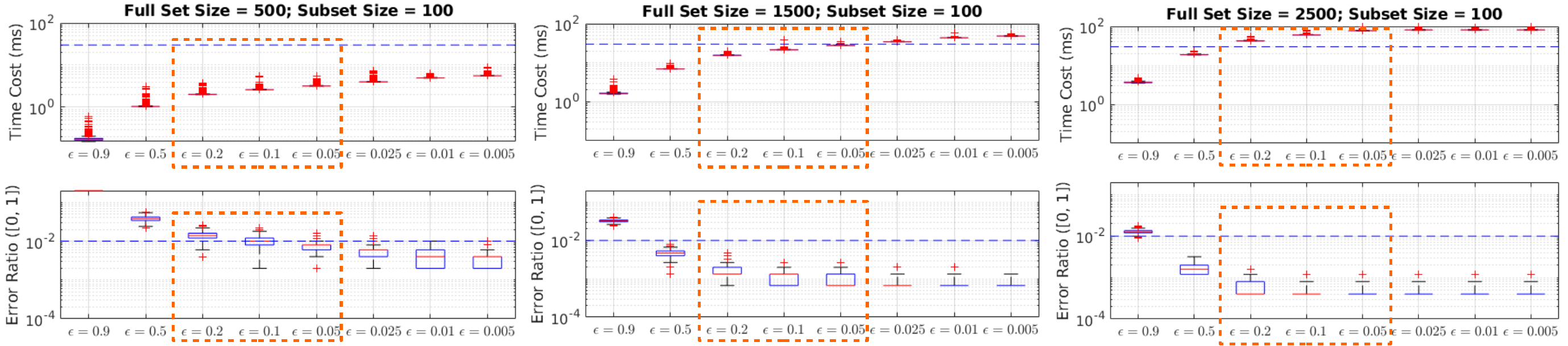}
    \caption{Lazier greedy with different decay factor $\epsilon$ under 3
    example configurations: selecting 100 features from full sets of 
    500, 1500, \& 2500.
	\textbf{First row}: time cost of lazier greedy.
	\textbf{Second row}: error ratio of lazier greedy.  
	For cross comparison, figures in both rows are with log-scale y-axis.  \label{fig:Simu_Lazier_Box}}
	\vspace{-0.5cm}
\end{figure*}

Fig. \ref{fig:Simu_Lazier_Surf} plots the simulation results for
computational time and error ratio as a function of sizes of the 
desired subset and the full set.  The error ratio uses the lazy-greedy 
outcome as the baseline, then computes the normalized RMS of the difference versus 
lazier greedy. 
The multiple surfaces for lazier-greedy correspond to different decay
factors $\epsilon$.  
Referring to the time cost graph, lazier greedy is 1-2 orders of magnitude
lower than greedy, depending on $\epsilon$.  The plot includes a
constant reference plane of $30 ms$ time cost (in blue). The preference is
to lie near to-or below-this reference plane, which lazier greedy can
achieve over large regions of its parameter space while lazy greedy cannot.
Moving to the error ratio graph, an error ratio of 0.01 indicates that the
subset selected with lazier greedy is less than 1\% different from the lazy
greedy baseline. Though slow, lazy greedy performs well for good
feature selection.  According to Fig. \ref{fig:Simu_Lazier_Surf}, the average
error ratio of lazier greedy is below $0.01$ for the majority of configuration
surfaces when $\epsilon \leq 0.1$.  
The graphs include three lines in the x-y visualization plane
corresponding to the target subset sizes used in the VSLAM experiments.
Lazier greedy with $\epsilon \leq 0.1$ consistently achieves a low error ratio,
yet consumes a fraction of time cost compared to lazy greedy.

To further identify an acceptable decay factor $\epsilon$, box-plots
of time cost and error ratio are presented in Fig. \ref{fig:Simu_Lazier_Box} 
under three configurations, which vary the number of matched features.  
We consider $\epsilon=0.1$ to be a favorable parameter
choice for good feature selection: the lazier greedy time cost is minimized
under the requirement of less-than-0.01 error ratio.  In what follows, all
experiments run lazier greedy with $\epsilon=0.1$.

\section{Good Feature Matching in VO/VSLAM Pipeline} \label{sec::goodmatch}
The prior discussion regarding the connection between tracked features
and {\em Max-logDet} subset selection led to an efficient good feature
selection algorithm.  Selection is based on the assumption that all 2D-3D
feature matchings are known, it only applies after data association (e.g.
similar to the existing works
\cite{zhang2015good,zhang2015optimally,carlone2019attention,zhao2018good2}). 
However, as shown in Figure \ref{fig:Illustration}, the time cost 
of data association occupies a significant portion of time (about 1/3) 
in the real-time pose tracking thread of feature-based VO/VSLAM.  To
reduce pose tracking time cost, consider translating the problem of good
feature selection to data association:
Given a set of 3D mapped features and a set of un-matched 2D measurements, 
only associate a constant-cardinality subset of 2D-3D matchings that will
minimize the error of the least squares pose optimization.  
In what follows, we discuss an efficient solution to the translate
problem, referred to as {\em good feature matching}. 

\subsection{Good Feature Matching in Monocular VO/VSLAM}
While the good feature selection problem applies to already associated data,
the good feature matching problem starts with a pool of 2D feature points and
3D map points whose correct associations are unknown.  The aim is to establish
which points should be targeted for matching and in what priority. 
Three modifications are made to transfer the previous solutions
to good feature matching, which is described in
Alg~\ref{alg:mono_goodMatch}: 

\textbf{(1)} Remove the dependency on 2D-3D matchings when constructing 
matrix $H_c$ (lines 1-4 of Alg~\ref{alg:mono_goodMatch}).  
Constructing $H_c$ requires knowing the covariance matrix of 2D measurement
$\Sigma_z(i)$ (as formulated in Eq~\eqref{eq:Residual_Formula}).
To avoid this information, assume a constant prior (e.g. a $2\times2$
identity matrix) at the initial stage of good feature matching (line 3 of
Alg~\ref{alg:mono_goodMatch});

\textbf{(2)} Add a feature matching condition check before updating the 
subsets in good feature selection (line 10 of Alg~\ref{alg:mono_goodMatch}): 
only when the current best 3D feature (with 
highest {\em logDet} margin gain) is successfully matched with some 2D 
measurement, should the matrix (feature) subset get updated accordingly.  
For the current best 3D feature, a search of possible 2D measurements 
is conducted on the image frame, with a size-fixed local search window 
(centered on the 2D projection of 3D feature).  If no 2D measurement can 
be matched to the current best 3D feature, it moves on to match the next 
best 3D feature (lines 15-17 of Alg~\ref{alg:mono_goodMatch}).  
By fusing good feature selection with feature matching, 
the selection strategy becomes an active matching algorithm: the feature 
matching effort prioritizes the subset with highest potential gain 
(in terms of {\em logDet}). 

\textbf{(3)} Information from successful feature matchings assists
follow-up good feature choice, by updating the measurement covariance 
$\Sigma_z(i)$ (and the associated block in $H_c(i)$) with measurement 
information (line 11 of Alg~\ref{alg:mono_goodMatch}).  
The measurement covariance $\Sigma_z(i)$ is assumed to be quadratic with 
the scale/pyramid level of measurement extraction; it is updated once the
pyramid level is known from 2D-3D feature matching.  The corrected block
$H_c(i)$ for the matched feature is concatenated into the selected submatrix,
and next iteration of greedy matching starts 
(lines 18-20 of Alg~\ref{alg:mono_goodMatch}).

\newcommand{\indSet}{\mathcal{I}}
\begin{algorithm}[!t]
 \KwData{ $P=\{p(1),\ p(2),\ ...\ ,\ p(n)\}$, $Z=\{z(1),\ z(2),\ ...\ ,\ z(m)\}$, $k$, $t_{max}$}
 \KwResult{ $M={<p(i), z(j)>},\ |M|=k$}
 
 $M\gets \emptyset,\,H_c^{sub}\gets \emptyset, 
   \,\indSet = \{1,\hdots,n\},\,t_{accu} = 0$\;
 \ForEach{$\text{ 3D feature }p(i)$}{			
 $\text{build Jacobians }H_x(i), \ H_p(i)$\;
 $W(i) = chol(\mathcal{I}_2+H_p(i)\Sigma_p(i)H_p(i)^T)$ \;
 $H_c(i) = W(i)^{-1} H_x(i)$ \;
 }
 
 \While{$|M|<k$ and $|\indSet|>0$ and ($t_{accu} < t_{max}$)}{								
 	$H_c^R \gets$ a random subset obtained by sampling
 	\hspace*{2.5em} $s=\frac{n}{k}\log(\frac{1}{\epsilon})$ non-repeated 
    random elements 
    \hspace*{2.5em} from $H_c$\;
 	\While{$|\indSet| > 0$ and ($t_{accu} < t_{max}$)}{
 		$i$ $ \gets \arg\max_{H_c(i) \in H_c^R} \log \det (H_c(i)^T H_c(i)$
 		\hspace*{12.5em} $+ [H_c^{sub}]^T [H_c^{sub}])$\;
 		\If{$\text{found matched measurement }z(j)\text{ for }p(i)$}{
 			$W(i) = chol(\Sigma_z(j)+H_p(i)\Sigma_p(i)H_p(i)^T)$ \;	
 			$H_c(i) = W(i)^{-1} H_x(i)$ \;
 			$M \gets M\cup <p(i), z(j)>$\;
 			break\;
 		}
 		\Else{									
 			$H_c^R \gets H_c^R \setminus H_c(i)$\;
 			$H_c^R \gets H_c^R \cup \text{a random sample from }H_c$\;
 		}
      $\indSet \gets \indSet \setminus\!\{i\}$
 	}	
	$H_c^{sub} \gets H_c^{sub}\cup H_c(i)$\;	
 	$H_c \gets H_c \setminus H_c(i)$\;			
 	$Z \gets Z \setminus\!z(j)$\;
 }
 \Return $M$.
 \caption{\small Good feature matching in mono VO/VSLAM. \label{alg:mono_goodMatch}}
\end{algorithm}

A fourth modification is made for SLAM problems.

\textbf{(4)} Rather than exhaustively search the candidate matching
pool for $k$ matches, 
the loops in lines 6 and 8 of Alg.~\ref{alg:mono_goodMatch}
include a time budget $t_{max}$ condition.  The time budget is sensible,
as the submodularity property is associated with diminishing returns
(i.e. the marginal value of the $j^{\text{th}}$ match is less than
earlier matches). Searching too long forfeits the task of balancing
accuracy and latency. In experiments $t_{max}=15$ms, and is rarely met.

\begin{algorithm}[!t]
 \KwData{ $P=\{p(1),\ ...\ ,\ p(n)\}$, $Z=\{z(1),\ ...\ ,\ z(m)\}$, $Z^r=\{z^r(1),\ ...\ ,\ z^r(s)\}$, $k$}
 \KwResult{ $M={<p(i), z(j), z^r(r)>},\ |M|=k$}
 \tcp{line 1-9 identical with monocular version}
 \setcounter{AlgoLine}{9}
 		\If{$\text{found matched left measurement }z(j)\text{ for }p(i)$}{
 			$W(i) = chol(\Sigma_z(j)+H_p(i)\Sigma_p(i)H_p(i)^T)$ \;
 			$H_c(i) = W(i)^{-1} H_x(i)$ \;
 			\If{$\text{found matched right measurement }z^r(d)\text{ for }p(i)$}{
 				$W^r(i) = chol(\Sigma_z^r(d)+H_p^r(i)\Sigma_p(i)H_p^r(i)^T)$ \;
 				$H_c(i) = [ H_c(i); W^r(i)^{-1} H_x^r(i) ]$ \;
 				$M \gets M\cup <p(i), z(j), z^r(d)>$\;
 			}
 			\Else{
 				$M \gets M\cup <p(i), z(j), \emptyset>$\;
 			}
 			break\;
 		}
 \tcp{rest of lines identical with line 15-22 of monocular version}
 \caption{\small Good feature matching in stereo VO/VSLAM. \label{alg:stereo_goodMatch}}
\end{algorithm}

\subsection{Good Feature Matching in Stereo VO/VSLAM}
Good feature matching applies to stereo cameras as well as to monocular
cameras.
Compared to monocular VO/VSLAM pipeline, stereo VO/VSLAM has an
additional module in data association: 
stereo matching, which associates measurements between left and right
frames.  
Since the stereo algorithm associates existing 3D mapped features to 2D 
measurements from both frames, each paired measurement provides twice
the number of rows to the least squares objective (in pose-only and
joint BA).  
Stereo methods also provide for instant initialization of
new map points through triangulated 2D measurements from the left and right 
frames.  
However, optimization for the current pose (as pursued in pose tracking)
only benefits from the stereo matchings associated with existing 3D
mapped features!  
By exploiting this property, we can design a lazy-stereo VO/VSLAM pipeline that 
has lower latency than the original stereo pipeline.
Stereo matching is postponed to after map-to-frame matching.  
Instead of searching for stereo matchings between all measurements, only
those measurements associated with 3D map points are matched. 
After pose optimization, the remaining measurements are 
stereo-matched and triangulated as new 3D mapped features. 

The lazy-stereo VO/VSLAM pipeline should have the same level of accuracy
\& robustness as the original pipeline, with reduced pose tracking latency.
Implementing the stereo good feature matching algorithm with the lazy-stereo
pipeline will further reduce latency while preserving accuracy \& robustness.  
Compared with the monocular Alg~\ref{alg:mono_goodMatch}, 
the stereo Alg~\ref{alg:stereo_goodMatch} has additional steps of stereo
matching at each successful iteration of map-to-frame feature
matching (line 13 of Alg~\ref{alg:stereo_goodMatch}). 
Depending on the matching outcome, the block $H_c(i)$ contains 
map-to-frame information only (no stereo matching found; 
line 11-12 of Alg~\ref{alg:stereo_goodMatch}), or 
both map-to-frame and left-to-right information (stereo matching found; 
line 14-15 of Alg~\ref{alg:stereo_goodMatch}). 

\subsection{Connection with Conventional Active Matching}
Conventionally, active matching is iteratively conducted with 2 major steps: 
1) a selecting step that chooses which 3D feature to match against, and 
2) a matching step that searches for best 2D measurements within a local
    area/window of image frame.  
The local area is typically refined during the active matching process,   as
more feature matches are found and used to improve the current camera pose estimate.  
For filter-based VO/VSLAM where the covariance matrix is easily assessable, 
refining the local search area on-the-fly is possible (by partially updating
the covariance matrix during the active matching process).  
Here we argue that for BA-based VO/VSLAM, refining local search area is not
necessary (and not efficient).  Instead, working with a fixed-size local
searching window is sufficient; it also improves the robustness towards
inconsistency and bias in state estimation.  Compared with the conventional
active matching, good feature matching also selects the best 3D feature to
match at each iteration, but the local search area for feature matching
remains fixed.

\section{Implementation and Evaluation} \label{sec::exp}
This section evaluates the performance of the proposed good feature
matching algorithm on a state-of-the-art feature-based
visual SLAM system, ORB-SLAM \cite{ORBSLAM}.  Applying the
proposed algorithms to the real-time tracking thread of ORB-SLAM 
(Alg~\ref{alg:mono_goodMatch} for monocular ORB-SLAM\footnote{https://github.com/raulmur/ORB\_SLAM} \& 
Alg~\ref{alg:stereo_goodMatch} for stereo ORB-SLAM2\footnote{https://github.com/raulmur/ORB\_SLAM2}), 
reduces the latency of pose tracking.  
Meanwhile, the tracking accuracy is either improved (on desktop) 
or the same as canonical ORB-SLAM (on low-power devices), and the robustness is
preserved (i.e. avoiding tracking failure).

ORB-SLAM involves two data association steps, 
keyframe-to-frame and map-to-frame.  Of the two, map-to-frame has the higher
time cost (see Fig. \ref{fig:Illustration}) and will always consist of points
with estimated 3D positions.  Thus we elect to incorporate good feature
matching into that module. 
Integrating the proposed good feature matching algorithm into 
the map-to-frame matching function of ORB-SLAM leads to several
changes, which provide additional, valuable run-time properties.
Since the keyframe-to-frame data association step will result in a set of
matches, $M_{K2F}$, the good feature matching process does not need to identify
a full set of $n_{GF}$ good feature matches. Instead it will identify a 
smaller set of $k = \textrm{min}(0, n_{GF} - |M_{K2F}|)$ good feature matches.
The modification has an additional advantage.
Figs.~\ref{fig:Simu_Lazier_Surf} and \ref{fig:Simu_Lazier_Box} indicate that the
time cost of lazier greedy grows past a given low threshold as the map size 
grows (e.g., the full set size).  Furthermore, the approximation error ratio 
grows as the subset size grows.  By limiting $k$ to a topping off 
functionality of $M_{K2F}$ that relates to the target cardinality $n_{GF}$, 
we are able to move the good feature matching implementation towards the 
lower subset sizes, therefore bounding the time cost and the error ratio.
Under Algs.~\ref{alg:mono_goodMatch} and \ref{alg:stereo_goodMatch},
the map-to-frame module prioritizes map point association according to the
{\em max-logDet} metric up to the target set size $k$, rather than attempt to 
match all visible local map points to the current measurements. 
This change establishes {\em when} to trigger active matching and
{\em how much} effort to apply (per the value $k$).
The follow-up pose tracking thread will utilize at most $n_{GF}$ associations,
which are faster to collect and to perform pose optimization with versus the
original implementation.

Due to the latency-reduction of good feature matching, there is typically
extra time between outputting the current pose estimate and receiving next
image. 
House cleaning and anticipatory calculations occur during this time.
House cleaning involves searching for additional map-to-frame feature
matchings, when the current frame is selected as a keyframe.  
The additional matches permit the local BA process to still take advantage of
the full set of feature matchings.  
Anticipatory calculations apply to the matrix preparation stage of good
feature matching, i.e. line 1-4 of Alg~\ref{alg:mono_goodMatch}.  
The steps are pre-computed to be immediately available for the next frame.  The
pre-computation further reduces the latency of good feature matching.  The
good-feature-matching enhanced ORB-SLAM is referred to as {\em GF-ORB-SLAM},
and {\em GF} for short.

For baseline comparison purposes, we integrate two reference methods into
ORB-SLAM that modify Algs.~\ref{alg:mono_goodMatch} and
\ref{alg:stereo_goodMatch} by prioritizing feature matching with simple
heuristics.  
One heuristic is purely-randomized matching, i.e., {\em Rnd} (no prioritization). 
The other heuristic prioritizes map points with a long tracking history since
they are more likely to be mapped accurately.  We refer to the second heuristic
method {\em Long}.

\subsection{Benchmarks}
The revised ORB-SLAM with good feature matching is evaluated against
available, state-of-the-art VO/VSLAM systems on four public
benchmarks:
\begin{enumerate}[wide, labelwidth=!]
  
  \item \textit{The EuRoC benchmark} \cite{burri2016euroc}, which contains 
    11 stereo-inertial sequences comprising 19 minutes of video,
    recorded in 3 different indoor environments.  Ground-truth tracks
    are provided using motion capture systems (Vicon \& Leica MS50).  We
    evaluate monocular (e.g. left camera) and stereo versions.
  \item\textit{The TUM-VI benchmark} \cite{schubert2018vidataset}, which 
    contains 28 stereo-inertial sequences of indoor and outdoor
    environments.  
    Only the 6 sequences (i.e. room1-6, in total 14 minutes of video)
    with complete coverage by MoCap ground truth are selected. 
    Compared with EuRoC, sequences in TUM-VI are recorded under much
    stronger camera motion, which is hard to track with monocular
    VO/VSLAM.  Only stereo methods are tested.
  \item \textit{The TUM-RGBD benchmark} \cite{sturm12iros_ws}, which is
    recorded with a Microsoft Kinect RGBD camera.  
    Three sequences that are relatively long (i.e. over 80 seconds each)
    and rich in camera motion are used in the evaluation.  The total
    length of videos selected is 5.5 minutes.  
    Compared with the previous two benchmarks, captured with global
    shutter cameras, the image quality of TUM-RGBD benchmark is
    lower, e.g.  rolling shutter, motion blur \cite{SVO2017}.  
    This benchmark tests monocular VO/VSLAM on low-end sensors and slow 
    motion, whereas the previous two test high-end sensors and fast motion.  
 \item \textit{The KITTI benchmark} \cite{KITTI}, which
   contains contains 11 stereo sequences recorded from a car in 
   urban and highway environments.  In total 40 minutes of video 
   are recorded, with individual recording duration ranging from 
   30 seconds to 8 minutes.  Ground truth tracks are provided by GPS/INS.  
   Unlike the earlier three indoor benchmarks, KITTI is a large-scale outdoor
   benchmark that characterizes self-driving applications.  Stereo VO/VSLAM
   methods are tested on KITTI.
\end{enumerate}

\subsection{Evaluation Metrics}
Since the focus of this work is on real-time pose tracking, 
all evaluations are performed on the instantaneous output of pose 
tracking thread; key-frame poses after posterior bundle adjustment 
are not used.  For fair comparison between VSLAM and VO methods, the loop 
closing modules are disabled in all ORB-SLAM variants.  
For the first three benchmarks that evaluate small-to-medium scale indoor
scenarios, \textit{absolute root-mean-square error} (RMSE) between ground 
truth track and SLAM estimated track is utilized as the accuracy metric 
(commonly used in SLAM evaluation 
\cite{sturm12iros_ws,nardi2015introducing,bodin2018slambench2,saeedi2018navigating}).  
For the last benchmark (KITTI outdoor), two relative metrics Relative Position 
Error (RPE) and Relative Orientation Error (ROE) are reported, 
as recommended \cite{KITTI}.  
Full evaluation results for both RMSE and RPE/ROE from all benchmarks are
provided externally%
\footnote{\url{https://github.com/ivalab/FullResults_GoodFeature}}%
\!\!.\,
Performance assessment involves a 10-run repeat for each configuration,
i.e., the benchmark sequence, the VO/VSLAM approach and the parameter 
(number of features tracked per frame).  
Results are reported if the VO/VSLAM approach works reliably under the
configuration; no tracking failure when running on a desktop, 
or at most 1 failure on a low-power device.  

Additional values recorded include the latency of real-time pose
tracking per frame, defined as the time span from receiving an image to
publishing the state estimate.
The latency of image capture and transmission are not included since
they are typically lower than that of VO/VSLAM algorithm, 
and are outside of the scope of this investigation.

This section first evaluates the accuracy-latency trade-off of
GF-ORB-SLAM against state-of-the-art monocular VO/VSLAM methods,  
then evaluates stereo version against stereo VO/VSLAM methods.  
In the process, we study the parameter-space of the {\em GF} modification in
order to identify the operational domain of any free parameters to fix them at
constant values in subsequent experiments.
The experiments are conducted on 3 desktops with identical 
configuration: Intel i7-7700K CPU (passmark score of 2581 per thread),
16 GB RAM, Ubuntu 14.04 and ROS Indigo environment. 
Finally, this section evaluates monocular GF-ORB-SLAM on low-power devices, 
suited for light-weight platforms such as micro aerial and
small ground vehicles.

\begin{figure}[!tb]
	\centering
	\includegraphics[width=0.95\linewidth]{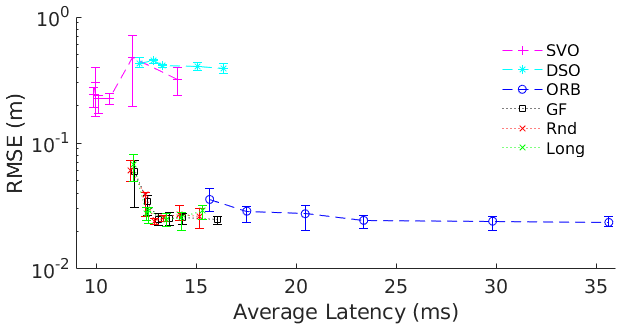} \\
	\includegraphics[width=0.95\linewidth]{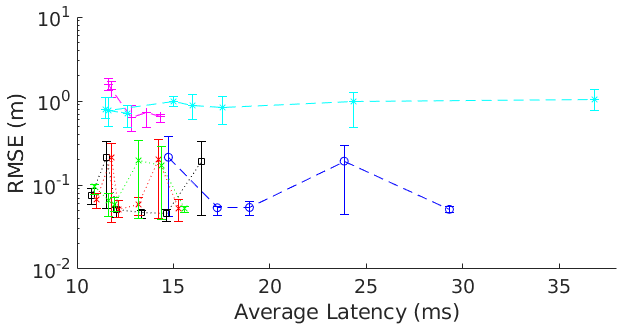} \\
	\includegraphics[width=0.95\linewidth]{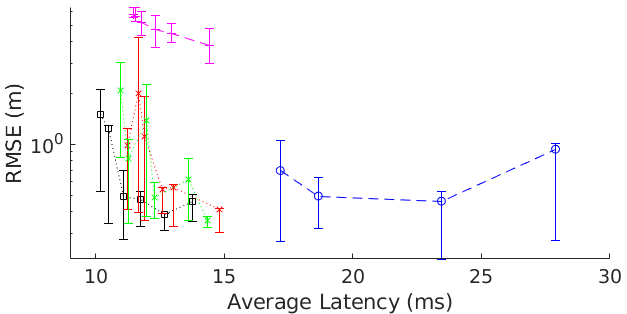} \\
	\caption{Latency vs. accuracy on 3 EuRoC Monocular sequences: 
      {\em MH 01 easy}, {\em V2 02 med}, and {\em MH 04 diff} (from top to bottom).  
	  Baseline systems are evaluated with {\em max feature number} ranging 
      from 150 to 2000; 
      ORB-SLAM variants are evaluated with {\em good feature number} ranging
      from 60 to 240, and {\em max feature number} fixed to 800.
      Only configurations with zero failure in a 10-run repeat are
      plotted (e.g. all configurations of DSO fail to track on {\em MH 04
      diff}, hence it is omitted in row 3).  
      The same rule applies subsequent latency vs. accuracy figures. 
      \label{fig:EuRoC_Mono_Time_vs_RMSE}}
      \vspace{-0.5cm}
\end{figure}

\begin{figure*}[!tb]
  \centering
  \includegraphics[width=0.4\linewidth]{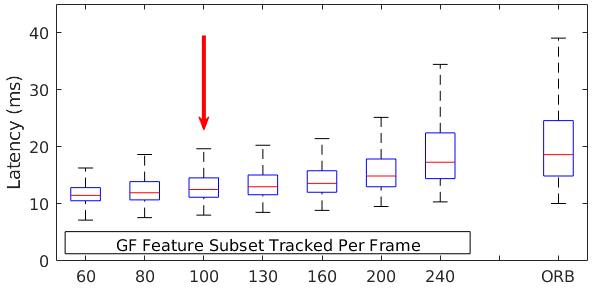} 
  \includegraphics[width=0.57\linewidth]{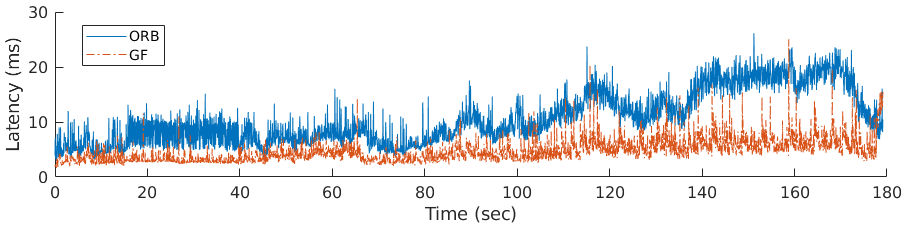} \\
  \caption{Latency vs. {\em good feature number} on EuRoC sequence 
    {\em MH 01 easy}. 
    \textbf{Left}: box-plots for {\em GF} and baseline {\em ORB}.  
    \textbf{Right}: the latency vs. time trend of {\em GF} under a 
    {\em good feature number} of 100 (marked with red arrow on left) and 
    {\em ORB} for 1 run.  \label{fig:EuRoC_Mono_Time_vs_Subset}}
    \vspace{-0.5cm}
\end{figure*}

\subsection{Latency vs. Accuracy: Mono VO/VSLAM}
In addition to the monocular ORB-SLAM baseline ({\em ORB}), two 
state-of-the-art monocular direct VO methods serve as baselines: 
{\em SVO}\footnote{\url{http://rpg.ifi.uzh.ch/svo2.html}} \cite{SVO2017} 
and 
{\em DSO}\footnote{\url{https://github.com/JakobEngel/dso}} \cite{DSO2017}.  
{\em SVO} is a light-weight direct VO system targeting low-latency pose
tracking while sacrificing tracking accuracy.  
The multi-threaded option in {\em SVO} is enabled, so that the depth
update/mapping runs on a separate thread from pose tracking.  
Compared with {\em SVO}, the direct objective in {\em DSO} is more comprehensive: 
it includes an explicit photometric model for image capturing.  
While {\em DSO} typically has better tracking performance than {\em SVO}, the
latency of {\em DSO} can be much higher (up to 3x).  
Unlike {\em ORB} and {\em SVO}, {\em DSO} has a single-thread
implementation only.  Its latency varies dramatically between regular
frames (e.g. $20$\,ms) and keyframes (e.g. $150$\,ms) \cite{DSO2017}.  
A multi-threaded {\em DSO} would only have the latency of regular frames, as
the keyframes can be processed in a separate thread.
Such a multi-threaded {\em DSO} is expected to be slightly less accurate than
the single-thread version, because the joint-optimized map points are no longer
immediately accessible for real-time pose tracking.  In our evaluation, an
idealized multi-threaded {\em DSO} is assumed: latency is evaluated on regular
frames only, while accuracy is evaluated on the single-thread {\em DSO}.

The latency and accuracy of VO/VSLAM systems can be adjusted through 
a few internal parameters.  
One key parameter that 
significantly impacts both latency and accuracy is the {\em max feature number},
i.e., the maximum number of features/patches tracked per frame.  
Running VO/VSLAM with high {\em max feature number} is beneficial for
accuracy \& robustness.  Meanwhile, lowering the {\em max feature
number} is preferred for latency reduction.  To evaluate the trade-off
between latency and accuracy for baseline systems ({\em ORB}, {\em SVO},
and {\em DSO}), all of them are configured to run 10-repeats for 
{\em max feature number} parameters ranging from 150 to 2000.  

For a given {\em max feature number}, ORB-SLAM latency can be reduced
via the proposed good feature matching algorithm.  Adjusting the 
{\em good feature number}, i.e. the number of good features being
matched in pose tracking, changes the latency.
Tests with the three ORB-SLAM variants ({\em GF}, {\em Rnd} and {\em Long}) 
are configured to run 10-repeat under {\em good feature number} values ranging 
from 60 to 240.  Meanwhile, the {\em max feature number} is fixed to
800, which yields a good balance of latency and accuracy for baseline 
{\em ORB}. 

\subsubsection{Parameter Exploration on EuRoC Monocular}
Fig.~\ref{fig:EuRoC_Mono_Time_vs_RMSE} presents the latency-accuracy
trade-off curves for monocular VO/VSLAM implementations on three example
EuRoC sequences 
%
Amongst the baseline methods, {\em ORB} has the best accuracy while 
{\em SVO} has the lowest latency.
Lowering the {\em max feature number} reduces the latency of {\em ORB} 
baseline, however, it comes with loss of tracking accuracy (e.g. the 1st blue
marker in row 2), or even the risk of track failure (e.g. the first 2 blue
markers are omitted in row 3).  
Meanwhile, a better latency-accuracy trade-off is achieved with the
proposed {\em GF} method.  
According to Fig. \ref{fig:EuRoC_Mono_Time_vs_RMSE}, the latency of {\em
GF} is in a similar range as {\em SVO}, but with the accuracy of {\em GF} 
being an order of magnitude better than both {\em SVO} and {\em DSO}.  
Furthermore, the accuracy-preserving property of {\em GF} is demonstrated 
when compared to the reference methods {\em Rnd} and {\em Long}.  
The latency-accuracy curves of {\em GF} are almost flat and lower than
the other two, once a reasonable number of good features are set to be
matched (e.g. starting from the 3rd black marker).  

The latency-reduction of {\em GF} is further illustrated in 
Fig. \ref{fig:EuRoC_Mono_Time_vs_Subset}, in which the 
{\em max feature number} is set to 800.  
Compared with {\em ORB}, the latency of {\em GF} has lower variance.
A good setting for the {\em good feature number} is 100, marked by 
a red arrow in Fig. \ref{fig:EuRoC_Mono_Time_vs_Subset}.  
The accuracy of {\em GF} with a {\em good feature number} of 100 is on par with
{\em ORB}, as quantified by the 3rd black marker in each row of Fig.
\ref{fig:EuRoC_Mono_Time_vs_RMSE}.  

\subsubsection{EuRoC Monocular}
Here, we report the accuracy \& latency of all monocular VO/VSLAM
methods under fixed configurations: 
the RMSE values are in Table~\ref{tab:EuRoC_Mono_RMSE}
(after a {\em Sim3} alignment to the ground truth), and 
the latency values in Table~\ref{tab:EuRoC_Mono_Latency}.  
For the three VO/VSLAM baselines, the {\em max feature number} is 800.
For the three {\em ORB} variants, the {\em max feature number} is 800
and the {\em good feature number} is 100.  
Results with any tracking failure are omitted from both tables.  
The {\em GF} subset selection does not impact the robustness of
ORB-SLAM: it works on all eight sequences that $ORB$ tracks.  
The average RMSE for all tracked sequences per method is given (i.e. All Avg.), 
as well as the average RMSE of the 5 sequences that all methods track 
successfully (i.e. Int. Avg.).

On each EuRoC sequence, the minimum RMSE is noted in bold.  
Interestingly, {\em GF} does not just preserve the accuracy \&
robustness of {\em ORB}; it further reduces the RMSE on several
sequences.  
On average, {\em GF} has the lowest RMSE over all evaluated VO/VSLAM methods.  
Furthermore, {\em GF} also has better overall accuracy when compared 
with two reference selection methods.  Though {\em Rnd} seems to have
lowest RMSE on multiple sequences, the margin between {\em Rnd} and {\em GF} 
small for them.  Meanwhile, both {\em Rnd} and {\em Long} lead to large
accuracy loss on the difficult sequence {\em MH 04 diff}, while {\em GF} 
improves RMSE.

According to Table~\ref{tab:EuRoC_Mono_Latency}, the average latency of 
{\em GF} is the lowest relative to all other methods: 
{\em GF} has an average latency 34\% lower than {\em ORB}.  
Compared with the direct methods, the latency of {\em GF} has lower
variance. The 1st quartile of {\em GF} latency is higher than direct methods, 
since feature extraction introduces a constant overhead.  
However, the 3rd quartile of {\em GF} latency is lower than direct
methods, which might occasionally spend too much time on direct
optimization.  

\begin{table}[tb!]
	\small
	\centering
	\caption{RMSE (m) on EuRoC Monocular Sequences}
	\begin{tabular}{|c|c|c|c|c|c|c|}
		\hline 
		\textbf{ } & 
		\multicolumn{6}{c|}{\bfseries \small VO/VSLAM} \\
		\textbf{\small Seq.} & SVO & DSO & ORB & GF & Rnd & Long \\
		\hline
		\textit{MH 01 easy}  & 0.227 & 0.407 & 0.027 & 0.025 & \textbf{0.024} & 0.029 	\\ 
		\textit{MH 02 easy}  & 0.761 & - & \textbf{0.034} & 0.043 & 0.038 & 0.040 		\\ 
		\textit{MH 03 med}   & 0.798 & 0.751 & 0.041 & 0.045 & 0.041 & \textbf{0.040} 	\\ 
		\textit{MH 04 diff}  & 4.757 & - & 0.699 & \textbf{0.492} & 1.110 & 1.377 		\\ 
		\textit{MH 05 diff}  & 3.505 & - & 0.346 & 0.464 & \textbf{0.216} & 0.915 		\\ 
		\textit{VR1 01 easy} & 0.726 & 0.950 & 0.057 & 0.037 & \textbf{0.036} & 0.037 	\\ 
		\textit{VR1 02 med}  & 0.808 & \textbf{0.536} &   -   &   -   &   -   & - 		\\ 
		\textit{VR1 03 diff} &  -    & - & - & - & - & - 								\\ 
		\textit{VR2 01 easy} & 0.277 & 0.297 & 0.025 & 0.024 & 0.025 & \textbf{0.023} 	\\ 
		\textit{VR2 02 med}  & 0.722 & 0.880 & 0.053 & \textbf{0.051} & \textbf{0.051} & 0.059 \\ 
		\textit{VR2 03 diff} &  -    & - & - & - & - & - \\ 
		\hline	
		\textbf{\small All Avg.} & 1.477 & 0.637 & 0.160 & \textbf{0.147} & 0.193 & 0.315 \\
		\textbf{\small Int. Avg.} & 0.550 & 0.657 & 0.041 & 0.036 & \textbf{0.035} & 0.038 \\
 	\hline	
	\end{tabular} 
	\label{tab:EuRoC_Mono_RMSE}
\end{table}

\begin{table}[tb!]
	\small
	\centering
	\caption{Latency (ms) on EuRoC Monocular Sequences}
	\begin{tabular}{|c|c|c|c|c|c|c|}
		\hline 
		\textbf{ } & 
		\multicolumn{6}{c|}{\bfseries \small VO/VSLAM}  \\
		\textbf{ } & SVO & DSO & ORB & GF & Rnd & Long \\
		\hline
		\small $\boldsymbol{Q_1}$    & 7.4  & \textbf{5.8}  & 13.9 & 10.3 & 10.0 & 10.0 \\
		\textbf{\small Avg.}      	 & 12.6 & 16.4 & 18.4 & \textbf{12.2} & 12.3 & 12.3 \\
		\small $\boldsymbol{Q_3}$    & 16.8 & 19.1 & 20.7 & 13.3 & 13.2 & \textbf{13.0} \\
		\hline	
	\end{tabular} 
	\label{tab:EuRoC_Mono_Latency}
\end{table}

\subsubsection{TUM-RGBD Monocular}
The RMSE values of all 6 methods (3 baseline VO/VSLAM, 3 {\em ORB} variants) 
evaluated on TUM-RGBD are summarized in Table~\ref{tab:TUM_RGBD_RMSE}.  
Due to the lower resolution, feature parameters used to obtain the
results are roughly half of those configured in EuRoC: 
{\em max feature number} of 400, and {\em good feature number} of 60 
(the lower limit recommended and tested in Fig.
\ref{fig:EuRoC_Mono_Time_vs_RMSE}, based on the trends in Figs.
\ref{fig:Simu_PnP_Res} and \ref{fig:EuRoC_Mono_Time_vs_RMSE}).
The average RMSE of {\em GF} is the 2nd lowest, next to the lowest RMSE
from {\em ORB}.  Not surprising, both the accuracy (e.g. average RMSE)
and the robustness (e.g. track failure) of direct methods are bad due to
rolling shutter effects.  

\begin{table}[tb!]
	\small
	\centering
	\caption{RMSE (m) on TUM-RGBD Sequences}
	\begin{tabular}{|c|c|c|c|c|c|c|}
		\hline 
		\textbf{ } & 
		\multicolumn{6}{c|}{\bfseries \small VO/VSLAM} \\
		\textbf{\small Seq.} & SVO & DSO & ORB & GF & Rnd & Long \\
		\hline
		\textit{f2 desk}        & 0.407 & 0.975 & \textbf{0.102} & 0.103 & 0.106 & 0.109 \\ 
 		\textit{f2 desk person} & 1.543 & - & \textbf{0.042} & 0.049 & 0.184 & 0.061 \\ 
		\textit{f3 long office} & - & 0.089 & 0.058 & 0.058 & \textbf{0.057} & 0.058 \\ 
		\hline	
		\textbf{\small All Avg.}& 0.975 & 0.532 & \textbf{0.067} & 0.070 & 0.116 & 0.076 \\ 
		\hline	
	\end{tabular} 
	\label{tab:TUM_RGBD_RMSE}
\end{table}

\begin{table}[tb!]
	\small
	\centering
	\caption{Latency (ms) on TUM-RGBD Sequences}
	\begin{tabular}{|c|c|c|c|c|c|c|}
		\hline 
		\textbf{ } & 
		\multicolumn{6}{c|}{\bfseries \small VO/VSLAM} \\
		\textbf{ } & SVO & DSO & ORB & GF & Rnd & Long \\
		\hline
		\small $\boldsymbol{Q_1}$    & 10.3 & \textbf{5.8}  & 8.3 & 7.1 & 6.8 & 6.8 \\
		\textbf{\small Avg.}  & 12.7 & 11.5 & 10.3 & 8.3 & 8.1 & \textbf{8.0} \\
		\small $\boldsymbol{Q_3}$    & 15.0 & 12.0 & 10.8 & 8.5 & 8.6 & \textbf{8.3} \\
		\hline	
	\end{tabular} 
	\label{tab:TUM_RGBD_Latency}
\end{table}

Latency reduction of {\em GF} is less significant than the previous EuRoC results: 
it saves around 19\% of average latency.  
Due to the lower image resolution and the relatively short duration of the
TUM-RGBD sequences, it is less likely to accumulate enough measurements towards
a large 3D feature map.  {\em GF} is best suited to localizing with a
relatively large-sized 3D map or domain; on a small map brute-force matching
will suffix.  This example demonstrates an example scenario with diminishing
returns.  However, for application scenarios that with improved image quality,
a larger domain of operation, and long-term duration, the advantage of {\em GF}
will be clearer.

\subsection{Latency vs. Accuracy: Stereo VO/VSLAM}
We also evaluate the latency-accuracy trade-off of 
stereo {\em GF} against state-of-the-art stereo VO/VSLAM systems. 
Compared to monocular VO/VSLAM, the amount of valid map points 
is much higher in stereo systems because of the extra stereo information.  
In the presence of a 3D map with high quality and quantity, the advantage 
of active map-to-frame matching is expected to be more significant 
than the monocular version.
The proposed good feature matching (Alg~\ref{alg:stereo_goodMatch}) 
is integrated into the sped-up ORB-SLAM, {\em Lz-ORB}.  In what 
follows, we again refer to the good feature enhanced ORB-SLAM as {\em GF}.  
As before, two heuristics are integrated into {\em Lz-ORB} as
reference methods, i.e. {\em Rnd} and {\em Long}.
Four baseline stereo systems are included in the evaluation as well: 
stereo {\em SVO}, stereo {\em DSO} (taken from published 
results \cite{wang2017stereo} on KITTI since no open-source 
implementation is available), canonical stereo ORB-SLAM ({\em ORB}), 
and {\em Lz-ORB}, a sped-up version of stereo ORB-SLAM based 
on the lazy-stereo pipeline described earlier.

The {\em max feature number} is adjusted for stereo baseline systems 
to obtain the trade-off curve between accuracy and latency.  
All 3 baseline systems ({\em SVO}, {\em ORB}, and {\em Lz-ORB}) are 
configured to have 10-repeat runs under {\em max feature number} ranging from
150 to 2000.  Meanwhile, the latency-accuracy trade-off of {\em GF} is obtained
by adjusting the {\em good feature number}, which is the the total number of
good features from both left and right frames that are matched to the local
map.  
All 3 ORB-SLAM variants ({\em GF}, {\em Rnd} and {\em Long}) are configured for
10-repeat runs under {\em good feature number} ranging from 60 to 240
(while {\em max feature number} is fixed). 

\subsubsection{Parameter Exploration for EuRoC Stereo}
\begin{figure}[!tb]
	\centering
	\includegraphics[width=0.95\linewidth]{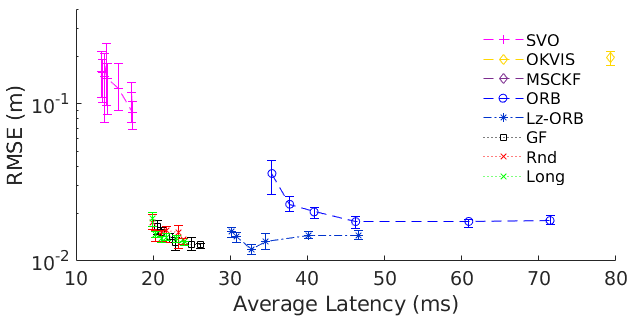} \\
	\includegraphics[width=0.95\linewidth]{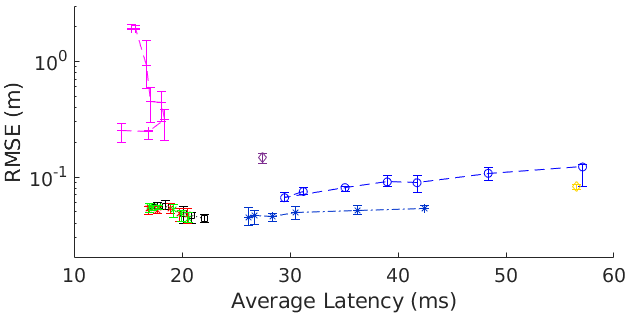} \\
	\includegraphics[width=0.95\linewidth]{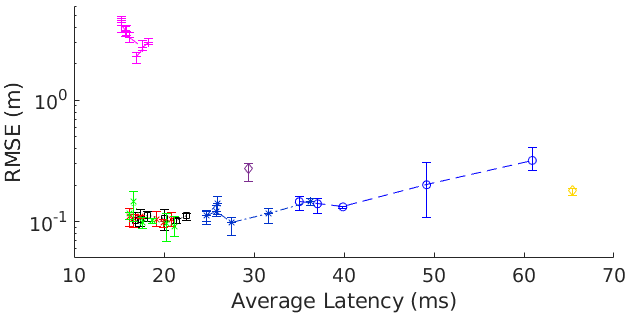} \\
	\caption{Latency vs. accuracy on 3 EuRoC Stereo sequences: {\em MH 01 easy}, {\em V2 02 med}, and {\em MH 04 diff} (from top to bottom). 
	Baseline systems are evaluated with {\em max feature number} ranging from 150 to 2000; 
	ORB-SLAM variants are evaluated with {\em good feature number} ranging from 60 to 240, and {\em max feature number} fixed to 800. \label{fig:EuRoC_Stereo_Time_vs_RMSE}}
	\vspace{-0.5cm}
\end{figure}

The latency-accuracy trade-off of stereo VO/VSLAM on three example EuRoC sequences 
can be found in Fig. \ref{fig:EuRoC_Stereo_Time_vs_RMSE}.  Among all 3 baseline 
systems, {\em Lz-ORB} has the best accuracy, while {\em SVO} has the lowest latency.  
Simply lowering the {\em max feature number} leads to accuracy drop or even 
track failure in {\em Lz-ORB}.  However, with {\em GF} the latency of pose 
tracking can be reduced to the same level as {\em SVO}, while the RMSE remains 
a magnitude lower than {\em SVO}.  
Two state-of-the-art stereo VINS systems, 
{\em OKVIS}\footnote{\url{https://github.com/ethz-asl/okvis}} \cite{leutenegger2015keyframe}
and {\em MSCKF}\footnote{\url{https://github.com/KumarRobotics/msckf_vio}} \cite{sun2018robust}, 
are evaluated as well.  Both VINS systems are assessed under the default parameters, therefore 
rather than having the full curve only one marker is presented in Fig. \ref{fig:EuRoC_Stereo_Time_vs_RMSE}.
The latency of {\em GF} is clearly lower than filter-based {\em MSCKF}, while the accuracy 
is even better than BA-based {\em OKVIS}. 
However, when comparing with two heuristics ({\em Rnd}, {\em Long}), 
the advantage of {\em GF} is harder to identify than monocular results. 

\begin{figure*}[!tb]
	\centering
	\includegraphics[width=0.4\linewidth]{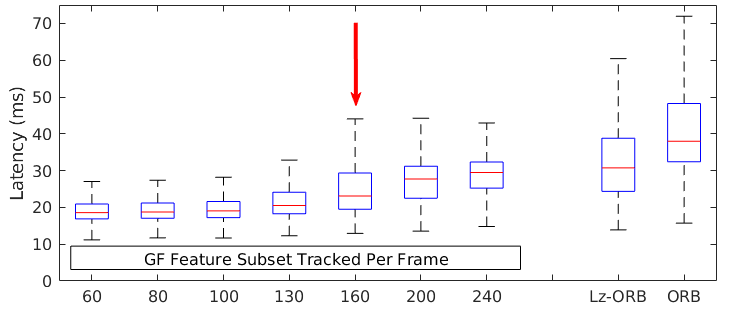}  
	\includegraphics[width=0.57\linewidth]{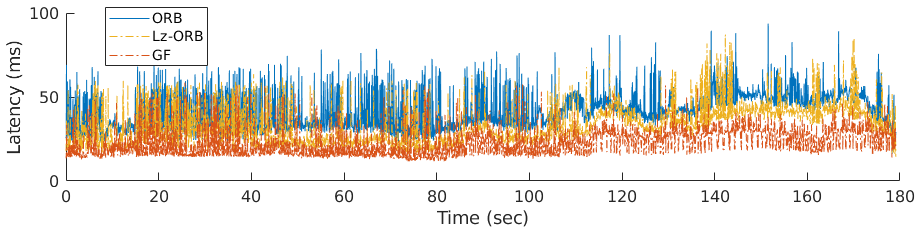} \\
	\caption{Latency vs. {\em good feature number} on EuRoC sequence {\em MH 01 easy}. \textbf{Left}: latency for {\em GF} under different {\em good feature number}, and 2 baselines {\em Lz-ORB} and {\em ORB}.  \textbf{Right}: the latency trend of {\em GF} under 160 {\em good feature number} (marked with red arrow at the left), {\em Lz-ORB} and {\em ORB} in 1 run.  \label{fig:EuRoC_Stereo_Time_vs_Subset}}
\end{figure*}

The latency reduction of {\em GF} is further illustrated in
Fig. \ref{fig:EuRoC_Stereo_Time_vs_Subset}.  
The {\em max feature number} being used in {\em Lz-ORB} and {\em ORB}
is 800, which balances accuracy and latency.  
Compared with the two non-GF baselines, the latency of {\em GF} is has a
lower upper bound.  A reasonable {\em good feature number} is 160,
since it yields low latency as well as high accuracy (the 3rd black mark
from the right in Fig. \ref{fig:EuRoC_Stereo_Time_vs_RMSE}).  

\subsubsection{EuRoC Stereo}
The RMSEs and latencies of all 6 stereo VO/VSLAM methods under the example configurations 
({\em max feature number} of 800 \& {\em good feature number} of 160) are summarized in 
Table~\ref{tab:EuRoC_Stereo_RMSE}.  
The results of 2 stereo VINS systems under default parameters are reported as well.  
Different from monocular VO/VSLAM, it is expected for stereo systems to estimate scale correctly.  
Therefore, each cell of Table~\ref{fig:EuRoC_Stereo_Time_vs_RMSE} reports 
the RMSE after {\em Sim3} alignment (as the 1st value) and the scale error
percentage (as the 2nd value).  
The lowest error within each category, i.e. VO/VSLAM or VINS, is highlighted in bold.  
Similar to the monocular experiment, {\em GF} is the lowest in terms of 
average RMSE and average scale error, compared with other stereo VO/VSLAM methods.  
Furthermore, the accuracy of {\em GF} is better than the two stereo VINS systems, 
while the robustness of {\em GF} is comparable to stereo VINS 
(each of them failed on 1 sequence). 
The advantage of {\em GF} over {\em Rnd} and {\em Long} can be 
verified as well: both {\em Rnd} and {\em Long} failed to track on 
{\em MH 02 easy} while {\em GF} succeed; the average RMSE and scale error of
{\em GF} are lower than the other two as well.  

The latency of all 8 stereo systems under the same configuration 
as Table~\ref{tab:EuRoC_LowPower_Latency} are summarized in 
Table~\ref{tab:EuRoC_Stereo_Latency}. 
The lowest latency is achieved with {\em SVO}, though its accuracy is an
order of magnitude higher than {\em GF}.  Additionally, its third quartile
latency is within 10\% of the {\em GF} third quartile latency.
The average latency reduction of {\em GF} is 
27.4\% when compared to {\em Lz-ORB}, and 
46.2\% when compared to {\em ORB}.  

\begin{table*}[tb!]
	\small
	\centering
	\caption{RMSE (m) and Scale Error (\%) on EuRoC Stereo Sequences}
	\begin{tabular}{|c|c|c|c|c|c|c||c|c|}
		\hline 
		\textbf{ } & 
		\multicolumn{6}{c|}{\bfseries \small VO/VSLAM} & 
		\multicolumn{2}{c|}{\bfseries \small VINS} \\
		\textbf{\small Seq.} & SVO & ORB & Lz-ORB & GF & Rnd & Long & OKVIS & MSCKF \\
		\hline
		\textit{MH 01 easy}  & 0.179 (0.6) & 0.021 (0.7) & \textbf{0.012} (\textbf{0.5}) & 0.013 (\textbf{0.5}) & 0.016 (\textbf{0.5}) & 0.014 (\textbf{0.5}) & \textbf{0.196} (\textbf{1.7}) & - \\ 
		\textit{MH 02 easy}  & - & 0.021 (0.3) & \textbf{0.018} (\textbf{0.1}) & 0.021 (\textbf{0.1}) & - & - & \textbf{0.114} (\textbf{1.4}) & 0.184 (2.0) \\  
		\textit{MH 03 med}   & 0.514 (2.3) & 0.029 (\textbf{0.3}) & \textbf{0.024} (0.4) & 0.025 (0.4) & 0.025 (0.4) & 0.025 (0.4) & \textbf{0.146} (\textbf{0.4}) & 0.260 (1.3) \\ 
		\textit{MH 04 diff}  & 3.753 (26.1) & 0.140 (1.1) & 0.120 (0.6) & 0.106 (\textbf{0.5}) & 0.104 (\textbf{0.5}) & \textbf{0.102} (0.6) & \textbf{0.179} (\textbf{0.9}) & 0.273 (1.0) \\ 
		\textit{MH 05 diff}  & 1.665 (4.9) & 0.096 (\textbf{0.2}) & \textbf{0.059} (\textbf{0.2}) & 0.068 (0.3) & 0.064 (\textbf{0.2}) & 0.103 (\textbf{0.2}) & \textbf{0.266} (\textbf{1.2}) & 0.356 (2.1) \\ 
		\textit{VR1 01 easy} & 0.264 (2.3) & \textbf{0.033} (0.8) & \textbf{0.033} (0.8) & 0.035 (\textbf{0.7}) & 0.035 (0.8) & 0.036 (\textbf{0.7}) & \textbf{0.046} (\textbf{0.4}) & 0.090 (0.9) \\ 
		\textit{VR1 02 med}  & 0.629 (11.2) & 0.064 (\textbf{0.4}) & 0.047 (0.7) & 0.038 (0.7) & \textbf{0.032} (0.7) & 0.036 (0.7) & \textbf{0.068} (0.5) & 0.123 (\textbf{0.3}) \\ 
		\textit{VR1 03 diff} & 0.655 (17.4) & 0.214 (\textbf{2.2}) & 0.112 (2.9) & \textbf{0.075} (\textbf{2.0}) & 0.080 (2.1) & 0.080 (2.1) & \textbf{0.120} (\textbf{1.0}) & 0.187 (1.1) \\ 
		\textit{VR2 01 easy} & 0.074 (1.7) & \textbf{0.031} (1.1) & 0.033 (0.9) & 0.044 (\textbf{0.5}) & 0.041 (0.6) & 0.042 (0.6) & \textbf{0.053} (0.8) & 0.071 (\textbf{0.3}) \\ 
		\textit{VR2 02 med}  & 0.447 (3.6) & 0.091 (\textbf{0.2}) & \textbf{0.046} (0.8) & 0.049 (0.9) & 0.053 (0.9) & 0.053 (0.9) & \textbf{0.083} (\textbf{0.7}) & 0.149 (1.0) \\ 
		\textit{VR2 03 diff} & \textbf{1.618} (\textbf{58.7}) & - & - & - & - & - & - & \textbf{1.162} (\textbf{39.9}) \\ 
		\hline	
		\textbf{\small All Avg.} & 0.980 (12.9) & 0.074 (0.7) & 0.050 (0.8) & \textbf{0.047} (\textbf{0.6}) & 0.050 (0.7) & 0.054 (0.8) & \textbf{0.127} (\textbf{0.9}) & 0.285 (5.0) \\ 
		\textbf{\small Int. Avg.} & 1.000 (8.7) & 0.087 (0.8) & 0.059 (0.9) & 0.055 (\textbf{0.7}) & \textbf{0.054} (0.8) & 0.060 (0.8) & \textbf{0.120} (\textbf{0.7}) & 0.189 (1.0) \\ 
		\hline	
	\end{tabular} 
	\label{tab:EuRoC_Stereo_RMSE}
\end{table*}
 
\begin{table*}[tb!]
	\small
	\centering
	\caption{Latency (ms) on EuRoC Stereo Sequences}
	\begin{tabular}{|c|c|c|c|c|c|c||c|c|}
		\hline 
		\textbf{ } & 
		\multicolumn{6}{c|}{\bfseries \small VO/VSLAM} & 
		\multicolumn{2}{c|}{\bfseries \small VINS} \\
		\textbf{ } & SVO & ORB & Lz-ORB & GF & Rnd & Long & OKVIS & MSCKF \\
		\hline
		\small $\boldsymbol{Q_1}$  & \textbf{8.6}  & 30.0 & 21.5 & 14.5 & 14.2 & 14.2 & 50.5 & \textbf{19.9} \\
		\textbf{\small Avg.}    & \textbf{16.4} & 38.5 & 28.5 & 20.7 & 19.9 & 20.1 & 65.1 & \textbf{28.3} \\
		\small $\boldsymbol{Q_3}$  & 23.3 & 44.2 & 32.1 & 24.2 & \textbf{22.5} & 22.9 & 80.3 & \textbf{36.0} \\
		\hline	
	\end{tabular} 
	\label{tab:EuRoC_Stereo_Latency}
\end{table*}

\begin{figure}[!tb]
	\centering
	\includegraphics[width=0.95\linewidth]{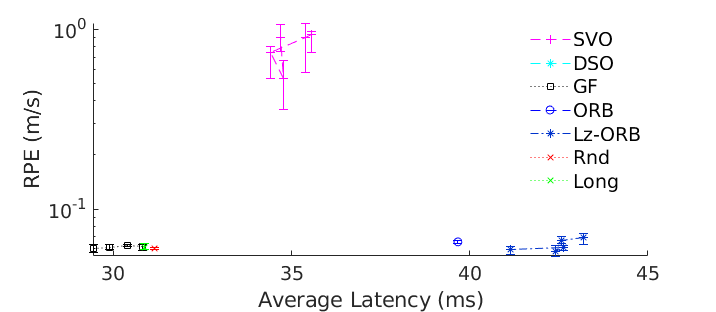} \\
	\includegraphics[width=0.95\linewidth]{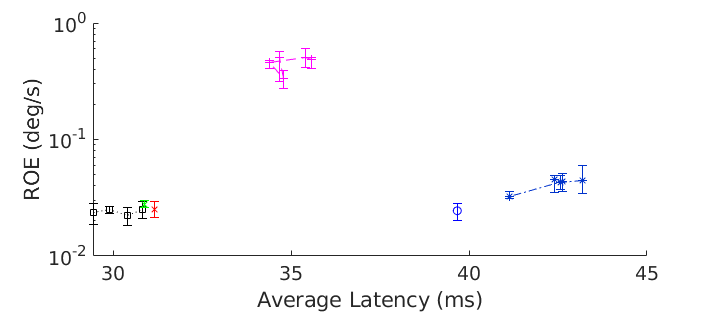} \\
	\caption{Latency vs. accuracy on KITTI sequence {\em 04}. }
	\label{fig:KITTI_Stereo_Time_vs_RE}
	\vspace{-0.5cm}
\end{figure}

\subsubsection{KITTI Stereo}
The latency-accuracy trade-off of stereo systems 
on a KITTI sequence {\em 04} are illustrated in 
Fig. \ref{fig:KITTI_Stereo_Time_vs_RE}.  Two relative 
metrics, RPE and ROE, are estimated with a sliding window 
of 100m.  As suggested in \cite{KITTI} and followed by \cite{ORBSLAM,wang2017stereo}, 
relative metrics are suited to evaluate accuracy of VO/VSLAM 
on outdoor large-scale sequences.  
Stereo {\em DSO} \cite{wang2017stereo} is not plotted, 
since there is no open-source implementation available 
to estimate tracking latency.  

%
\begin{table*}[tb!]
	\small
	\centering
	\caption{RPE (m/s), and ROE (deg/s) on KITTI Stereo Sequences}
	\begin{tabular}{|c|c|c|c|c|c|c|c|}
		\hline 
		\textbf{ } & 
		\multicolumn{7}{c|}{\bfseries \small VO/VSLAM} \\
		\textbf{\small Seq.} & SVO & DSO & ORB & Lz-ORB & GF & Rnd & Long \\
		\hline
		\textit{00}  & 0.632, 0.898 & \textbf{0.140, 0.163} & 0.143, 0.169 & 0.144, 0.169 	& 0.142, 0.167 & 0.143, 0.168 & 0.145, 0.170 \\ 
		\textit{01}  & 4.418, 1.585 & 0.236, \textbf{0.041} & \textbf{0.234}, 0.068 & - 	& - & - & - \\ 
		\textit{02}  & 0.553, 0.683 & 0.107, \textbf{0.053} & 0.106, 0.061 & - 			& \textbf{0.105}, 0.062 & \textbf{0.105}, 0.062 & \textbf{0.105}, 0.062 \\ 
		\textit{03}  & 0.208, 0.170 & 0.061, \textbf{0.030} & 0.057, 0.032 & 0.057, 0.038	& \textbf{0.054}, 0.036 & - & - \\ 
		\textit{04}  & 0.534, 0.337 & \textbf{0.059, 0.024} & 0.066, 0.024 & 0.061, 0.044	& 0.061, 0.025 & - & - \\ 
		\textit{05}  & 0.312, 0.229 & 0.047, \textbf{0.042} & \textbf{0.045}, 0.048 & 0.046, 0.048 	& \textbf{0.045}, 0.047 & \textbf{0.045}, 0.047 & \textbf{0.045}, 0.047 \\ 
		\textit{06}  & 0.879, 1.537 & \textbf{0.061, 0.051} & 0.067, 0.053 & 0.066, \textbf{0.051}	& 0.065, 0.055 & - & - \\ 
		\textit{07}  & 0.244, 0.326 & \textbf{0.048, 0.052} & 0.049, 0.057 & 0.050, 0.059 	& \textbf{0.048}, 0.060 & - & - \\ 
		\textit{08}  & 0.456, 0.304 & \textbf{0.225, 0.055} & 0.226, 0.061 & - 			& \textbf{0.225}, 0.063 & 0.226, 0.063 & 0.226, 0.064 \\ 
		\textit{09}  & 0.494, 0.491 & 0.070, \textbf{0.054} & 0.065, 0.058 & -			& \textbf{0.062}, 0.058 & - & - \\ 
		\textit{10}  & 0.668, 0.874 & \textbf{0.062, 0.043} & \textbf{0.062}, 0.050 & -	& \textbf{0.062}, 0.052 & - & - \\ 
		\hline	
		\textbf{\small All Avg.} & 0.854, 0.676 & 0.101, \textbf{0.055} & 0.102, 0.062 & \textbf{0.071}, 0.068 & 0.087, 0.062 & 0.130, 0.085 & 0.130, 0.086 \\ 
		\hline	
	\end{tabular} 
	\label{tab:KITTI_Stereo_RE}
\end{table*}

\begin{table}[tb!]
	\small
	\centering
	\caption{Latency (ms) on KITTI Stereo Sequences}
	\begin{tabular}{|c|c|c|c|c|c|c|c|}
		\hline 
		\textbf{ } & 
		\multicolumn{7}{c|}{\bfseries \small VO/VSLAM} \\
		\textbf{\small Seq.} & SVO & DSO & ORB & Lz-ORB & GF & Rnd & Long \\
		\hline
		\small $\boldsymbol{Q_1}$  & 26.0 & - & 33.9 & 26.4 & 21.6 & \textbf{21.3} & 21.4 \\
		\textbf{\small Avg.}       & 34.7 & - & 44.8 & 43.7 & 29.4 & \textbf{29.1} & 29.4 \\
		\small $\boldsymbol{Q_3}$  & 40.7 & - & 54.0 & 60.6 & 31.6 & \textbf{31.2} & 31.4 \\
		\hline	
	\end{tabular} 
	\label{tab:KITTI_Stereo_Latency}
	\vspace{-0.5cm}
\end{table}

Similar to the previous results on indoor scenarios, 
{\em GF} is at the bottom-left of the latency-accuracy plane.  
The latency of {\em GF} is lower than {\em ORB}, while the relative error of
{\em GF} is at the same level as {\em ORB}.
{\em GF} also behaves more robustly than the two reference heuristics: both
{\em Rnd} and {\em Long} tracks on one out of four configurations, while {\em
GF} works on all four configurations (i.e. has four black-square markers).

Furthermore, we report the RPE \& ROE of 7 stereo systems 
in Table~\ref{tab:KITTI_Stereo_RE}, and the latency in 
Table~\ref{tab:KITTI_Stereo_Latency}.  Since the image 
resolution in KITTI is double that of previous benchmarks 
(captured by VGA/WVGA cameras), the numbers in Table~\ref{tab:KITTI_Stereo_RE} 
and \ref{tab:KITTI_Stereo_Latency} are collected under 
{\em max feature number} of 1500.  To be consistent with 
EuRoC stereo results, the {\em good feature number} is also 
fixed to 160.  Stereo {\em DSO} results are obtained 
from the authors' online site%
\footnote{\url{https://vision.in.tum.de/research/vslam/stereo-dso}}%
\!\!\!\!,\ \ \ 
from a single run of each sequence.  All other methods 
are evaluated under 10-repeat runs. 

According to Table~\ref{tab:KITTI_Stereo_RE}, {\em GF} and {\em ORB}
track 10 out of 11 sequences, with {\em GF} having a lower RPE than {\em ORB}.  
The two reference methods, {\em Rnd} and {\em Long}, failed to track three
sequences.  The performance of direct systems varies: {\em SVO} has the worst 
accuracy on all 11 sequences, while stereo {\em DSO} works slightly better than
{\em GF} in terms of accuracy and robustness.  
Several reasons contribute to the performance of {\em DSO}: the motion profile
of a car is smoother than that of a MAV or hand-held camera; revisits happen at a
lower rate than indoor scenarios; and the lighting condition is well-controlled
with few low-light cases.

Latency reduction of {\em GF} is illustrated in 
Table~\ref{tab:KITTI_Stereo_Latency}.  On average {\em GF} 
has 30\% less latency than {\em ORB} and {\em Lz-ORB} to track 
a pair of stereo images.  {\em GF} also has a much lower upper 
bound of pose tracking latency.  The latency of {\em SVO} is 
higher than {\em GF}. The latency of {\em DSO} is not available.

\subsubsection{TUM-VI Stereo}

Under a {\em max feature number} of 600 and a {\em good feature number} of 
160, we report the RMSE, scale error, and latency of all 8 stereo
systems, in Tables~\ref{tab:TUM_Stereo_RMSE} and \ref{tab:TUM_Stereo_Latency}. 
Compared to stereo VINS systems {\em OKVIS} and {\em MSCKF}, 
{\em GF} is less robust (i.e. failed to track on \textit{room3}).  
We further argue that the drop in robustness of {\em GF} is not due to 
good feature matching: the original {\em ORB} tracks all 6 sequences, while our
vanilla implementation of {\em Lz-ORB} fails on {\em room3} (and all 3 variants
thereafter).  
The track failure for {\em room3} should be resolved  
with a better {\em Lz-ORB} implementation or with the incorporation of 
IMU measurements.

%
\begin{table*}[tb!]
	\small
	\centering
	\caption{RMSE (m) and Scale Error (\%) on TUM-VI Stereo Sequences}
	\begin{tabular}{|c|c|c|c|c|c|c||c|c|}
		\hline 
		\textbf{ } & 
		\multicolumn{6}{c|}{\bfseries \small VO/VSLAM} & 
		\multicolumn{2}{c|}{\bfseries \small VINS} \\
		\textbf{\small Seq.} & SVO & ORB & Lz-ORB & GF & Rnd & Long & OKVIS & MSCKF \\
		\hline
		\textit{room1}  & 1.036 (95.9) 	& 0.290 (8.0) & 0.057 (\textbf{1.3}) & 0.048 (1.6) & \textbf{0.040} (1.4) & 0.044 (1.4) & \textbf{0.065 (0.6)} & 0.152 (0.8) \\ 
		\textit{room2}  & 1.208 (97.9) 	& 0.412 (11.4)& 0.191 (2.7) & \textbf{0.141 (1.8)} & 0.145 (1.9) & \textbf{0.141 (1.8)} & \textbf{0.101 (0.8)} & 0.148 (1.5) \\ 
		\textit{room3}  & 1.204 (84.2) 	& 0.160 (4.0) & - 			& - & - & - & \textbf{0.057 (0.4)} & 0.201 (2.4) \\ 
		\textit{room4}  & - 			& 0.156 (4.0) & 0.036 (\textbf{0.8}) & 0.035 (1.0) & 0.035 (1.0) & \textbf{0.034} (1.0) & \textbf{0.026 (0.3)} & 0.130 (1.8) \\ 
		\textit{room5}  & - 			& 0.349 (11.7)& 0.028 (\textbf{0.4}) & 0.029 (\textbf{0.3}) & 0.029 (\textbf{0.3}) & \textbf{0.028} (0.4) & \textbf{0.048 (0.3)} & 0.137 (2.1) \\ 
		\textit{room6}  & 0.756 (55.6) 	& 0.039 (3.3) & 0.031 (\textbf{1.5}) & 0.032 (\textbf{1.5}) & \textbf{0.030 (1.5)} & \textbf{0.030} (1.6) & \textbf{0.038 (0.7)} & 0.116 (1.2) \\ 
		\hline	
		\textbf{\small All Avg.}  & 1.051 (83.4) & 0.234 (7.1) & 0.069 (1.3) & 0.057 (1.3) & \textbf{0.056 (1.2)} & \textbf{0.056} (1.3) & \textbf{0.056 (0.5)} & 0.147 (1.7) \\
		\textbf{\small Int. Avg.} & 1.000 (83.1) & 0.247 (7.7) & 0.093 (1.8) & 0.074 (\textbf{1.6}) & \textbf{0.072 (1.6)} & \textbf{0.072 (1.6)} & \textbf{0.068 (0.7)} & 0.139 (1.2) \\
		\hline	
	\end{tabular} 
	\label{tab:TUM_Stereo_RMSE}
\end{table*}

\begin{table*}[tb!]
	\small
	\centering
	\caption{Latency (ms) on TUM-VI Stereo Sequences}
	\begin{tabular}{|c|c|c|c|c|c|c||c|c|}
		\hline 
		\textbf{ } & 
		\multicolumn{6}{c|}{\bfseries \small VO/VSLAM} & 
		\multicolumn{2}{c|}{\bfseries \small VINS} \\
		\textbf{ } & SVO & ORB & Lz-ORB & GF & Rnd & Long & OKVIS & MSCKF \\
		\hline
		\small $\boldsymbol{Q_1}$  & 12.2 & 23.2 & 17.7 & 12.3 & \textbf{11.6} & \textbf{11.6} & \textbf{5.8} & 15.1 \\
		\textbf{\small Avg.}       & 18.1 & 29.3 & 22.9 & 15.1 & 14.4 & \textbf{14.3} & \textbf{11.5} & 22.2 \\
		\small $\boldsymbol{Q_3}$  & 22.3 & 34.0 & 26.6 & 16.2 & 15.7 & \textbf{15.4} & \textbf{17.0} & 28.5 \\
		\hline	
	\end{tabular} 
	\label{tab:TUM_Stereo_Latency}
\end{table*}

For the 5 sequences where {\em GF} succeeds, the RMSE of {\em GF} is 
lower than the vision-only baselines ({\em SVO}, {\em ORB} and {\em Lz-ORB}).
The average RMSE of vision-only {\em GF} on 5 tracking sequences 
is close to that of visual-inertial {\em OKVIS}, while being lower than that of
{\em MSCKF}.  Furthermore, {\em GF} leads to a 40.2\% reduction of average
latency versus {\em Lz-ORB}, and 53.2\% latency reduction versus {\em ORB}, 
according to Table~\ref{tab:TUM_Stereo_Latency}.  

Interestingly the average RMSE of the two heuristics ({\em Rnd} and {\em Long}) 
are slightly lower than for {\em GF} and they have a lower latency than {\em GF}. 
The advantage of {\em Rnd} and {\em Long} is largely due to the set-up of
TUM-VI room sequences: these sequences are captured in a small room, with the
camera performing repeated circular motion.  
In such a set-up, the 3D map of the entire room gets constructed after one to two 
circles, with high quality and a high quantity of features.  The success rate
of map-to-frame feature matching will be high for a small-scale world with frequent
revisits.  Under these conditions, simple heuristics such as 
{\em Rnd} and {\em Long} provide sufficient feature matching inliers for pose
tracking with less computation demands than search methods such as {\em GF}.

Based on the performance guarantees described in Section~\ref{sec::lazier}, 
which are in expectation, there may exist situations where lazier greedy will
operate with similar performance to randomized methods.  The TUM scenarios
highlight one such set of situations. However, outside of these situations,
the two methods are anticipated to diverge. 
The poorer pose estimation of {\em Rnd} will then affect the long term pose
tracking performance due to the recursive estimation nature of SLAM.
This assertion is supported by the evaluation results in 
EuRoC (medium-scale SLAM) and KITTI (large-scale SLAM), 
where {\em GF} has better accuracy and robustness than {\em Rnd}.

\subsection{Real-time Tracking on Low-Power Devices}
Here, the proposed {\em GF} modification is deployed on three low-power devices
with limited processing capabilities, which typically serve as on-board
processing units for light weight platforms.  The low-power devices tested
include:

\noindent
1) X200CA: a light-weight laptop with an Intel Pentium 2117U processor 
(Passmark score: 1662 per thread) and 4 GB of RAM.  The processor has 
2 cores and consumes 17W. \\
2) Jetson TX2: a 64-bit embedded single-board computer system, 
containing a hybrid processing unit (2 Denver2 + 4 ARM A57) 
and 8 GB of RAM.  Power consumption is 7.5W. \\
3) Euclid: a 64-bit embedded single-board computer system, 
with a Intel Atom x7-Z8700 processor (Passmark score: 552 
per thread) and 4 GB of RAM.  The processor has 4 cores 
and consumes 4W.  

{\em GF}, and three other monocular VO/VSLAM baselines, are 
deployed and evaluated with EuRoC monocular sequences.
To run {\em ORB} variants near real-time, the pyramid levels for ORB
feature extraction were reduced to 3 from 8, and the {\em max feature number} 
set to 400.  As a consequence, the robustness performance of the {\em ORB} 
variants is worse than the previous EuRoC Mono results.  
In what follows, we relax the robustness condition slightly, and report 
results with 1 tracking failure in 10 runs as well (marked with underline).  

\begin{table*}[tb!]
	\small
	\centering
	\caption{RMSE (m) On EuRoC Monocular Systems, Running on Low-power Devices.}
	\begin{tabular}{|c|c|c|c|c||c|c|c||c|c|c|c|c|c|}
		\hline 
		\textbf{ } & 
		\multicolumn{4}{c|}{\bfseries \small X200CA} & 
		\multicolumn{3}{c|}{\bfseries \small Jetson} & 
		\multicolumn{6}{c|}{\bfseries \small Euclid} \\
		\textbf{\small Seq.} & SVO & DSO & ORB & GF &
		DSO & ORB & GF &
		SVO & DSO & ORB & GF & SVOMSF & VIMono \\
		\hline
		\textit{MH 01 easy}  & 0.327 & - & \underline{0.041} & \textbf{0.036}     & - & \textbf{0.033} & 0.037              & \underline{0.244} & - & 0.044 & \textbf{0.041} & 0.29 & 0.20 \\ 
		\textit{MH 02 easy}  & - & - & 0.053 & \textbf{0.047}         & - & \textbf{0.046} & \underline{0.135}             & - & - & \textbf{0.044} & 0.045 & 0.31 & 0.18 \\ 
		\textit{MH 03 med}   & 1.14 & - & \textbf{0.050} & 0.056     & - & \textbf{0.055} & 0.059              & 1.21 & - & \underline{\textbf{0.050}} & \underline{0.051} & 0.66 & 0.17 \\ 
		\textit{MH 04 diff}  & - & - & \textbf{0.281} & 0.457         & - & \underline{\textbf{0.231}} & -                 & - & - & 0.232 & \underline{0.248} & 2.02 & \textbf{0.12} \\ 
		\textit{MH 05 diff}  & \underline{2.54} & - & 0.289 & \textbf{0.233}    & - & \textbf{0.258} & 0.340              & \underline{2.84} & - & - & \underline{\textbf{0.158}} & 0.87 & 0.35 \\  
		\textit{VR1 01 easy} & 0.552 & - & \textbf{0.036} & \textbf{0.036}     & 0.826 & \textbf{0.036} & \textbf{0.036} & 0.645 & - & \textbf{0.036} & 0.040 & 0.36 & 0.05 \\ 
		\textit{VR1 02 med}  & \textbf{0.730} & - & - & - & - & - & - & \textbf{0.857} & - & - & - & 0.78 & \textbf{0.12} \\ 
		\textit{VR1 03 diff} & - & - & - & - & - & - & - & - & - & - & - & - & \textbf{0.10} \\ 
		\textit{VR2 01 easy} & 0.397 & 0.295 & 0.032 & \textbf{0.029} & 0.288 & 0.029 & \textbf{0.027}          & 0.402 & 0.300 & \textbf{0.030} & 0.032 & 0.33 & 0.08 \\ 
		\textit{VR2 02 med}  & \underline{0.634} & 0.832 & - & \textbf{0.213}    & \textbf{0.941} & - & -                  & 0.688 & - & - & - & 0.59 & \textbf{0.08} \\ 
		\textit{VR2 03 diff} & - & - & - & - & - & - & - & - & - & - & - & - & \textbf{0.17} \\ 
		\hline	
		\textbf{\small All Avg.} & 0.903 & 0.564 & \textbf{0.112} & 0.138 & 0.685 & \textbf{0.098} & 0.106 & 0.984 & 0.300 & \textbf{0.073} & 0.088 & 0.69 & 0.15 \\
		\textbf{\small Int. Avg.} & - & - & \textbf{0.112} & 0.128 & - & \textbf{0.076} & 0.106 & - & - & \textbf{0.073} & 0.076 & 0.66 & 0.13 \\
		\hline	
	\end{tabular} 
	\label{tab:EuRoC_LowPower_RMSE}
\end{table*}

\begin{table*}[tb!]
	\small
	\centering
	\caption{Latency (ms) On EuRoC Monocular Systems, Running on Low-power Devices.}
	\begin{tabular}{|c|c|c|c|c||c|c|c||c|c|c|c|c|c|}
		\hline 
		\textbf{ } & 
		\multicolumn{4}{c|}{\bfseries \small X200CA} & 
		\multicolumn{3}{c|}{\bfseries \small Jetson} & 
		\multicolumn{6}{c|}{\bfseries \small Euclid} \\
		\textbf{ } & 
		SVO & DSO & ORB & GF &
		DSO & ORB & GF &
		SVO & DSO & ORB & GF & SVOMSF & VIMono \\
		\hline
		\small $\boldsymbol{Q_1}$& \textbf{8.6}  & 12.4 & 19.3 & 14.5 & \textbf{21.5} & 30.2 & 25.7 & \textbf{12.6} & 21.5 & 28.7 & 24.8 & 29.8 & 88.9 \\
		\textbf{\small Avg.} 	 & \textbf{9.8}  & 15.0 & 24.6 & 18.7 & 32.1 & 35.1 & \textbf{31.1} & \textbf{13.4} & 37.3 & 35.9 & 32.6 & 37.5 & 153.9 \\
		\small $\boldsymbol{Q_3}$& \textbf{12.5} & 16.4 & 28.5 & 21.0 & 37.5 & 38.8 & \textbf{33.1} & \textbf{15.9} & 51.1 & 41.4 & 39.3 & 42.1 & 209.5 \\
		\hline	
	\end{tabular} 
	\label{tab:EuRoC_LowPower_Latency}
	\vspace{-0.5cm}
\end{table*}

The RMSEs on all three low-power devices are summarized in 
Table~\ref{tab:EuRoC_LowPower_RMSE}, while the latencies are 
summarized in Table~\ref{tab:EuRoC_LowPower_Latency}.  
The {\em good feature number} is set to 60 given 
the {\em max feature number} of 400 (similar to the TUM RGBD benchmark case).

\noindent
1) When running on X200CA, {\em GF} has the 2nd lowest average RMSE 
(23\% higher than {\em ORB}).  However, the robustness of {\em GF} 
is slightly better than {\em ORB} and {\em SVO}: it tracks on 8 
sequences without failure, while the other 2 baselines track 7 
sequences and with failure.  When comparing on the 7 sequences 
that {\em ORB} tracks, {\em GF} only introduces 14\% to average RMSE.  
The strength of {\em SVO} is the low-latency; though 
the average latency of {\em GF} is 24\% less than {\em ORB}, it 
is almost twice that of {\em SVO}.  \\
2) The released binary of {\em SVO} does not support 64-bit Jetson TX2, 
therefore only 3 methods are assessed on Jetson.  Similar to the X200CA
results, {\em GF} is slightly worse than {\em ORB} in terms of average 
RMSE (by 8\%).  Notice {\em GF} is also less robust than {\em ORB}, 
as it introduces additional tracking failure on sequences {\em MH 02 easy} 
and {\em MH 04 diff}.  The latency reduction of {\em GF} is also small: 
11\% less than {\em ORB}.  \\
3) When running on Euclid, {\em GF} introduces 20\% more error in terms 
of average RMSE.  Again, notice that {\em GF} works on {\em MH 05 diff} 
while {\em ORB} cannot.  If we only take the 6 sequences that {\em ORB} 
tracks into account, {\em GF} only introduces 4\% to average RMSE.  
However, the latency reduction of {\em GF} for Euclid is smaller than the
Jetson results: only 9\% time savings.
Apart from the 4 monocular VO/VSLAM systems, we also include 
the VINS results \cite{delmerico2018benchmark} evaluated on a UP Board, 
which has almost identical hardware specifications as Euclid.  
The RMSE of the VINS methods, labeled {\em SVOMSF}\cite{delmerico2018benchmark} 
and {\em VIMono}\cite{qin2018vins}, are obtained by {\em Sim3} alignment 
to ground truth, which is identical with our evaluation.  
With additional input from inertial sensors, VINS
are clearly more robust than vision-only systems.  However, the accuracy
of VINS is poorer than vision-only ones (when scale corrected).  
Furthermore, the latency of the VINS approaches is much higher than
vision-only systems, which suggests the scalability of VINS is also poor
for low-power devices.  Therefore, for VO/VSLAM and VINS, a combination of 
algorithm improvements (e.g. Good Feature) and hardware improvements 
may be required to achieve low latency and good accuracy on embedded devices.

\begin{figure}[!tb]
	\centering
	\includegraphics[width=\linewidth]{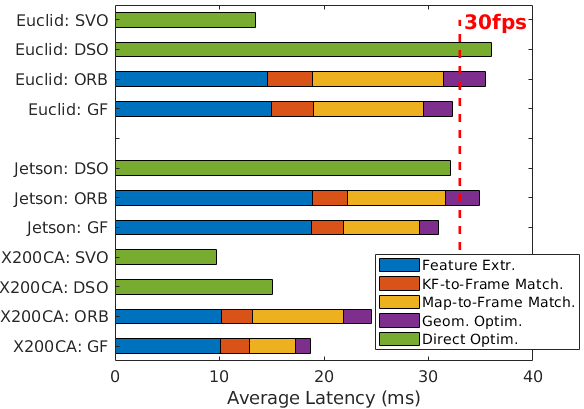} \\
	\caption{Latency breakdown for all modules in pose tracking pipeline, running on low-power devices. \label{fig:Bar_Lowpower}}
	\vspace{-0.5cm}
\end{figure}

When the computate resources (e.g. processor speed, cache size) are highly
limited, the latency reduction of {\em GF} is less significant.  
Preservation of accuracy \& robustness, on the other hand, scales relatively
well on different devices (only with slight drop).  The limited scalability to
devices such as Jetson \& Euclid is mostly due to the sequential nature of the
proposed {\em GF} algorithm.  As embedded device hardware specifications
improve, in terms of compute power and core quantity, we anticipate that
improvements will favor the {\em GF} variant (as demonstrated on desktop and
X200CA).  Even on current embedded platforms, the small amount of latency
reduced by {\em GF} could be important: it turns the near real-time {\em ORB}
into a real-time applicable VSLAM system, as illustrated 
in Fig. \ref{fig:Bar_Lowpower}.

Given the time cost of feature extraction (Fig.
\ref{fig:Bar_Lowpower}), efforts to move feature extraction onto FPGA
devices \cite{fang2017fpga, quigley2018open} are crucial.  
The times in Fig. \ref{fig:Bar_Lowpower}
reflect a coarser pyramid and smaller feature extraction numbers. 
Off-loading the original ORB-SLAM pyramidal feature extraction block 
to an FPGA will have significant savings and would help preserve the 
accuracy properties of the original front end.
When combining FPGA off-loading with the Good Feature matching method 
(and possibly also IMU integration), we expect the performance-efficiency of 
VSLAM on low-power devices be similar to the desktop outcomes 
(e.g. Table \ref{tab:EuRoC_Mono_RMSE}).


\section{Conclusion} \label{sec::conc}
This paper presents an active map-to-frame feature matching 
method, good feature matching, which reduces the computational cost 
(and therefore latency) of VO/VSLAM, while preserving the 
accuracy and robustness of pose tracking.  Feature matching 
is connected to the submatrix selection problem.  To that 
end, the {\em Max-logDet} matrix revealing metric was shown 
to perform best via simulated scenarios.  
For application to active feature matching, the combination of
deterministic selection (greedy) and randomized acceleration (random
sampling) is studied.  The proposed good feature matching algorithm is
integrated into monocular and stereo feature-based VSLAM systems, followed by
evaluation on multiple benchmarks and computate platforms.  
Good feature matching is shown to be an efficiency enhancement for low-latency
VO/VSLAM, while preserving, if not improving, the accuracy and robustness of
VO/VSLAM.  
Though the focus of this paper is reducing the latency 
of VO/VSLAM, the idea of active \& logDet-guided feature matching 
is general: it can be extended to other feature modules (e.g. 
line features \cite{GoodLine}) or localization tasks (e.g. 
image-based localization \cite{zhao2019maphash}).




%



%

\addtolength{\textheight}{-3cm}
\bibliographystyle{IEEEtran}
\bibliography{./full_references}

\begin{thebibliography}{10}
\providecommand{\url}[1]{#1}
\csname url@samestyle\endcsname
\providecommand{\newblock}{\relax}
\providecommand{\bibinfo}[2]{#2}
\providecommand{\BIBentrySTDinterwordspacing}{\spaceskip=0pt\relax}
\providecommand{\BIBentryALTinterwordstretchfactor}{4}
\providecommand{\BIBentryALTinterwordspacing}{\spaceskip=\fontdimen2\font plus
\BIBentryALTinterwordstretchfactor\fontdimen3\font minus
  \fontdimen4\font\relax}
\providecommand{\BIBforeignlanguage}[2]{{%
\expandafter\ifx\csname l@#1\endcsname\relax
\typeout{** WARNING: IEEEtran.bst: No hyphenation pattern has been}%
\typeout{** loaded for the language `#1'. Using the pattern for}%
\typeout{** the default language instead.}%
\else
\language=\csname l@#1\endcsname
\fi
#2}}
\providecommand{\BIBdecl}{\relax}
\BIBdecl

\bibitem{MonoSLAM}
A.~J. Davison, I.~D. Reid, N.~D. Molton, and O.~Stasse, ``{M}ono{SLAM}:
  Real-time single camera {SLAM},'' \emph{IEEE Transactions on Pattern Analysis
  and Machine Intelligence}, vol.~29, no.~6, pp. 1052--1067, 2007.

\bibitem{PTAM}
G.~Klein and D.~Murray, ``Parallel tracking and mapping for small {AR}
  workspaces,'' in \emph{IEEE/ACM International Symposium on Mixed and
  Augmented Reality}, 2007, pp. 225--234.

\bibitem{ORBSLAM}
R.~Mur-Artal, J.~M.~M. Montiel, and J.~D. Tardos, ``{ORB}-{SLAM}: a versatile
  and accurate monocular {SLAM} system,'' \emph{IEEE Transactions on Robotics},
  vol.~31, no.~5, pp. 1147--1163, 2015.

\bibitem{engel2014lsd}
J.~Engel, T.~Sch{\"o}ps, and D.~Cremers, ``{LSD}-{SLAM}: {L}arge-{S}cale
  {D}irect monocular {SLAM},'' in \emph{European Conference on Computer
  Vision}.\hskip 1em plus 0.5em minus 0.4em\relax Springer, 2014, pp. 834--849.

\bibitem{SVO2017}
C.~Forster, Z.~Zhang, M.~Gassner, M.~Werlberger, and D.~Scaramuzza, ``S{VO}:
  Semidirect visual odometry for monocular and multicamera systems,''
  \emph{IEEE Transactions on Robotics}, vol.~33, no.~2, pp. 249--265, 2017.

\bibitem{DSO2017}
J.~Engel, V.~Koltun, and D.~Cremers, ``Direct sparse odometry,'' \emph{IEEE
  Transactions on Pattern Analysis and Machine Intelligence}, vol.~40, no.~3,
  pp. 611--625, 2017.

\bibitem{newcombe2011kinectfusion}
R.~A. Newcombe, S.~Izadi, O.~Hilliges, D.~Molyneaux, D.~Kim, A.~J. Davison,
  P.~Kohi, J.~Shotton, S.~Hodges, and A.~Fitzgibbon, ``{K}inect{F}usion:
  Real-time dense surface mapping and tracking,'' in \emph{IEEE International
  Symposium on Mixed and Augmented Reality}, 2011, pp. 127--136.

\bibitem{whelan2016elasticfusion}
T.~Whelan, R.~F. Salas-Moreno, B.~Glocker, A.~J. Davison, and S.~Leutenegger,
  ``Elastic{F}usion: Real-time dense {SLAM} and light source estimation,''
  \emph{The International Journal of Robotics Research}, vol.~35, no.~14, pp.
  1697--1716, 2016.

\bibitem{fang2017fpga}
W.~Fang, Y.~Zhang, B.~Yu, and S.~Liu, ``{FPGA}-based {ORB} feature extraction
  for real-time visual {SLAM},'' in \emph{IEEE International Conference on
  Field Programmable Technology}, 2017, pp. 275--278.

\bibitem{quigley2018open}
M.~Quigley, K.~Mohta, S.~S. Shivakumar, M.~Watterson, Y.~Mulgaonkar,
  M.~Arguedas, K.~Sun, S.~Liu, B.~Pfrommer, V.~Kumar \emph{et~al.}, ``The open
  vision computer: An integrated sensing and compute system for mobile
  robots,'' \emph{arXiv preprint arXiv:1809.07674}, 2018.

\bibitem{zhang2017visual}
Z.~Zhang, A.~A. Suleiman, L.~Carlone, V.~Sze, and S.~Karaman, ``Visual-inertial
  odometry on chip: An algorithm-and-hardware co-design approach,'' 2017.

\bibitem{kueng2016low}
B.~Kueng, E.~Mueggler, G.~Gallego, and D.~Scaramuzza, ``Low-latency visual
  odometry using event-based feature tracks,'' in \emph{IEEE/RSJ International
  Conference on Intelligent Robots and Systems}, 2016, pp. 16--23.

\bibitem{rebecq2017real}
H.~Rebecq, T.~Horstschaefer, and D.~Scaramuzza, ``Real-time visualinertial
  odometry for event cameras using keyframe-based nonlinear optimization,'' in
  \emph{British Machine Vision Conference}, vol.~3, 2017.

\bibitem{zhu2017event}
A.~Z. Zhu, N.~Atanasov, and K.~Daniilidis, ``Event-based visual inertial
  odometry.'' in \emph{IEEE Conference on Computer Vision and Pattern
  Recognition}, 2017, pp. 5816--5824.

\bibitem{calonder2010brief}
M.~Calonder, V.~Lepetit, C.~Strecha, and P.~Fua, ``B{RIEF}: Binary robust
  independent elementary features,'' in \emph{European Conference on Computer
  Vision}.\hskip 1em plus 0.5em minus 0.4em\relax Springer, 2010, pp. 778--792.

\bibitem{leutenegger2011brisk}
S.~Leutenegger, M.~Chli, and R.~Y. Siegwart, ``B{RISK}: Binary robust invariant
  scalable keypoints,'' in \emph{IEEE International Conference on Computer
  Vision}, 2011, pp. 2548--2555.

\bibitem{rublee2011orb}
E.~Rublee, V.~Rabaud, K.~Konolige, and G.~Bradski, ``O{RB}: An efficient
  alternative to {SIFT} or {SURF},'' in \emph{IEEE International Conference on
  Computer Vision}, 2011, pp. 2564--2571.

\bibitem{vandergheynst2012freak}
P.~Vandergheynst, R.~Ortiz, and A.~Alahi, ``F{REAK}: Fast retina keypoint,'' in
  \emph{IEEE Conference on Computer Vision and Pattern Recognition}, 2012, pp.
  510--517.

\bibitem{leutenegger2015keyframe}
S.~Leutenegger, S.~Lynen, M.~Bosse, R.~Siegwart, and P.~Furgale,
  ``Keyframe-based visual--inertial odometry using nonlinear optimization,''
  \emph{The International Journal of Robotics Research}, vol.~34, no.~3, pp.
  314--334, 2015.

\bibitem{pvribyl2016camera}
B.~P{\v{r}}ibyl, P.~Zem{\v{c}}{\'\i}k, and M.~{\v{C}}ad{\'\i}k, ``Camera pose
  estimation from lines using {P}l\"ucker coordinates,'' in \emph{British
  Machine Vision Conference}, 2015, pp. 1--12.

\bibitem{PLSLAM}
R.~Gomez-Ojeda, F.-A. Moreno, D.~Zu{\~n}iga-No{\"e}l, D.~Scaramuzza, and
  J.~Gonzalez-Jimenez, ``{PL}-{SLAM}: a stereo {SLAM} system through the
  combination of points and line segments,'' \emph{IEEE Transactions on
  Robotics}, 2019.

\bibitem{GoodLine}
Y.~Zhao and P.~A. Vela, ``Good line cutting: towards accurate pose tracking of
  line-assisted {VO/VSLAM},'' in \emph{European Conference on Computer
  Vision}.\hskip 1em plus 0.5em minus 0.4em\relax Springer, 2018, pp. 516--531.

\bibitem{yang2018challenges}
N.~Yang, R.~Wang, X.~Gao, and D.~Cremers, ``Challenges in monocular visual
  odometry: Photometric calibration, motion bias, and rolling shutter effect,''
  \emph{IEEE Robotics and Automation Letters}, vol.~3, no.~4, pp. 2878--2885,
  2018.

\bibitem{schubert2018direct}
D.~Schubert, N.~Demmel, V.~Usenko, J.~Stuckler, and D.~Cremers, ``Direct sparse
  odometry with rolling shutter,'' in \emph{European Conference on Computer
  Vision}.\hskip 1em plus 0.5em minus 0.4em\relax Springer, 2018, pp. 682--697.

\bibitem{burri2016euroc}
M.~Burri, J.~Nikolic, P.~Gohl, T.~Schneider, J.~Rehder, S.~Omari, M.~W.
  Achtelik, and R.~Siegwart, ``The {EuRoC} micro aerial vehicle datasets,''
  \emph{The International Journal of Robotics Research}, vol.~35, no.~10, pp.
  1157--1163, 2016.

\bibitem{zhao2018good2}
Y.~Zhao and P.~Vela, ``Good feature selection for least squares pose
  optimization in {VO/VSLAM},'' in \emph{IEEE/RSJ International Conference on
  Intelligent Robots and Systems}, 2018, pp. 3569--3574.

\bibitem{JCBB}
J.~Neira and J.~D. Tard{\'o}s, ``Data association in stochastic mapping using
  the joint compatibility test,'' \emph{IEEE Transactions on Robotics and
  Automation}, vol.~17, no.~6, pp. 890--897, 2001.

\bibitem{KalmanSAC}
A.~Vedaldi, H.~Jin, P.~Favaro, and S.~Soatto, ``Kalman{SAC}: Robust filtering
  by consensus,'' in \emph{IEEE International Conference on Computer Vision},
  2005, pp. 633--640.

\bibitem{1pRANSAC}
J.~Civera, O.~G. Grasa, A.~J. Davison, and J.~Montiel, ``1-{P}oint {RANSAC} for
  extended {K}alman filtering: Application to real-time structure from motion
  and visual odometry,'' \emph{Journal of Field Robotics}, vol.~27, no.~5, pp.
  609--631, 2010.

\bibitem{cvivsic2015stereo}
I.~Cvi{\v{s}}i{\'c} and I.~Petrovi{\'c}, ``Stereo odometry based on careful
  feature selection and tracking,'' in \emph{European Conference on Mobile
  Robots}, 2015, pp. 1--6.

\bibitem{zhang2015optimally}
G.~Zhang and P.~A. Vela, ``Optimally observable and minimal cardinality
  monocular {SLAM},'' in \emph{IEEE International Conference on Robotics and
  Automation}, 2015, pp. 5211--5218.

\bibitem{zhang2015good}
G.~Zhang and P.~A. Vela, ``Good features to track for visual {SLAM},'' in
  \emph{IEEE Conference on Computer Vision and Pattern Recognition}, 2015, pp.
  1373--1382.

\bibitem{GoodFeaturesToTrack}
J.~Shi and C.~Tomasi, ``Good features to track,'' in \emph{IEEE Conference on
  Computer Vision and Pattern Recognition}, 1994, pp. 593--600.

\bibitem{sala2006landmark}
P.~Sala, R.~Sim, A.~Shokoufandeh, and S.~Dickinson, ``Landmark selection for
  vision-based navigation,'' \emph{IEEE Transactions on Robotics}, vol.~22,
  no.~2, pp. 334--349, 2006.

\bibitem{shi2013feature}
Z.~Shi, Z.~Liu, X.~Wu, and W.~Xu, ``Feature selection for reliable data
  association in visual {SLAM},'' \emph{Machine Vision and Applications}, pp.
  1--16, 2013.

\bibitem{covarianceRecovery}
M.~Kaess and F.~Dellaert, ``Covariance recovery from a square root information
  matrix for data association,'' \emph{Robotics and Autonomous Systems},
  vol.~57, no.~12, pp. 1198--1210, 2009.

\bibitem{zhang2005entropy}
S.~Zhang, L.~Xie, and M.~D. Adams, ``Entropy based feature selection scheme for
  real time simultaneous localization and map building,'' in \emph{IEEE/RSJ
  International Conference on Intelligent Robots and Systems}, 2005, pp.
  1175--1180.

\bibitem{lerner2007landmark}
R.~Lerner, E.~Rivlin, and I.~Shimshoni, ``Landmark selection for task-oriented
  navigation,'' \emph{IEEE Transactions on Robotics}, vol.~23, no.~3, pp.
  494--505, 2007.

\bibitem{cheein2009feature}
F.~A. Cheein, G.~Scaglia, F.~di~Sciasio, and R.~Carelli, ``Feature selection
  criteria for real time {EKF}-{SLAM} algorithm,'' \emph{International Journal
  of Advanced Robotic Systems}, vol.~6, no.~3, p.~21, 2009.

\bibitem{carlone2019attention}
L.~Carlone and S.~Karaman, ``Attention and anticipation in fast visual-inertial
  navigation,'' \emph{IEEE Transactions on Robotics}, vol.~35, no.~1, pp.
  1--20, 2019.

\bibitem{ila2017fast}
V.~Ila, L.~Polok, M.~Solony, and K.~Istenic, ``Fast incremental bundle
  adjustment with covariance recovery,'' in \emph{IEEE International Conference
  on 3D Vision}, 2017, pp. 175--184.

\bibitem{ila2017slam++}
V.~Ila, L.~Polok, M.~Solony, and P.~Svoboda, ``{SLAM}++-a highly efficient and
  temporally scalable incremental {SLAM} framework,'' \emph{The International
  Journal of Robotics Research}, vol.~36, no.~2, pp. 210--230, 2017.

\bibitem{gu1996efficient}
M.~Gu and S.~C. Eisenstat, ``Efficient algorithms for computing a strong
  rank-revealing {QR} factorization,'' \emph{SIAM Journal on Scientific
  Computing}, vol.~17, no.~4, pp. 848--869, 1996.

\bibitem{boutsidis2009improved}
C.~Boutsidis, M.~W. Mahoney, and P.~Drineas, ``An improved approximation
  algorithm for the column subset selection problem,'' in \emph{ACM-SIAM
  Symposium on Discrete Algorithms}, 2009, pp. 968--977.

\bibitem{shamaiah2010greedy}
M.~Shamaiah, S.~Banerjee, and H.~Vikalo, ``Greedy sensor selection: Leveraging
  submodularity,'' in \emph{IEEE Conference on Decision and Control}, 2010, pp.
  2572--2577.

\bibitem{jawaid2015submodularity}
S.~T. Jawaid and S.~L. Smith, ``Submodularity and greedy algorithms in sensor
  scheduling for linear dynamical systems,'' \emph{Automatica}, vol.~61, pp.
  282--288, 2015.

\bibitem{summers2016submodularity}
T.~H. Summers, F.~L. Cortesi, and J.~Lygeros, ``On submodularity and
  controllability in complex dynamical networks,'' \emph{IEEE Transactions on
  Control of Network Systems}, vol.~3, no.~1, pp. 91--101, 2016.

\bibitem{drineas2008relative}
P.~Drineas, M.~W. Mahoney, and S.~Muthukrishnan, ``Relative-error {CUR} matrix
  decompositions,'' \emph{SIAM Journal on Matrix Analysis and Applications},
  vol.~30, no.~2, pp. 844--881, 2008.

\bibitem{drineas2012fast}
P.~Drineas, M.~Magdon-Ismail, M.~W. Mahoney, and D.~P. Woodruff, ``Fast
  approximation of matrix coherence and statistical leverage,'' \emph{Journal
  of Machine Learning Research}, vol.~13, no. Dec, pp. 3475--3506, 2012.

\bibitem{boutsidis2014near}
C.~Boutsidis, P.~Drineas, and M.~Magdon-Ismail, ``Near-optimal column-based
  matrix reconstruction,'' \emph{SIAM Journal on Computing}, vol.~43, no.~2,
  pp. 687--717, 2014.

\bibitem{mirzasoleiman2015lazier}
B.~Mirzasoleiman, A.~Badanidiyuru, A.~Karbasi, J.~Vondr{\'a}k, and A.~Krause,
  ``Lazier than lazy greedy.'' in \emph{AAAI Conference on Artificial
  Intelligence}, 2015, pp. 1812--1818.

\bibitem{hassidim2017robust}
A.~Hassidim and Y.~Singer, ``Robust guarantees of stochastic greedy
  algorithms,'' in \emph{International Conference on Machine Learning}, 2017,
  pp. 1424--1432.

\bibitem{activeSearch}
A.~Davison, ``Active search for real-time vision,'' in \emph{IEEE International
  Conference on Computer Vision}, vol.~1, 2005, pp. 66--73.

\bibitem{activeMatching}
M.~Chli and A.~J. Davison, ``Active matching,'' in \emph{European Conference on
  Computer Vision}.\hskip 1em plus 0.5em minus 0.4em\relax Springer, 2008, pp.
  72--85.

\bibitem{scalableActiveMatching}
A.~Handa, M.~Chli, H.~Strasdat, and A.~Davison, ``Scalable active matching,''
  in \emph{IEEE Conference on Computer Vision and Pattern Recognition}, 2010,
  pp. 1546--1553.

\bibitem{davison2018futuremapping}
A.~J. Davison, ``{F}uture{M}apping: The computational structure of spatial {AI}
  systems,'' \emph{arXiv preprint arXiv:1803.11288}, 2018.

\bibitem{LandmarkImpact}
J.~Sola, T.~Vidal-Calleja, J.~Civera, and J.~M.~M. Montiel, ``Impact of
  landmark parametrization on monocular {EKF}-{SLAM} with points and lines,''
  \emph{International Journal of Computer Vision}, vol.~97, no.~3, pp.
  339--368, 2012.

\bibitem{golub2012matrix}
G.~H. Golub and C.~F. Van~Loan, \emph{Matrix computations}.\hskip 1em plus
  0.5em minus 0.4em\relax JHU Press, 2012, vol.~3.

\bibitem{minoux1978accelerated}
M.~Minoux, ``Accelerated greedy algorithms for maximizing submodular set
  functions,'' in \emph{Optimization techniques}.\hskip 1em plus 0.5em minus
  0.4em\relax Springer, 1978, pp. 234--243.

\bibitem{horn1990matrix}
R.~A. Horn, R.~A. Horn, and C.~R. Johnson, \emph{Matrix analysis}.\hskip 1em
  plus 0.5em minus 0.4em\relax Cambridge university press, 1990.

\bibitem{schubert2018vidataset}
D.~Schubert, T.~Goll, N.~Demmel, V.~Usenko, J.~Stueckler, and D.~Cremers, ``The
  {TUM VI} benchmark for evaluating visual-inertial odometry,'' in
  \emph{IEEE/RJS International Conference on Intelligent Robot Systems}, 2018.

\bibitem{sturm12iros_ws}
J.~Sturm, W.~Burgard, and D.~Cremers, ``Evaluating egomotion and
  structure-from-motion approaches using the {TUM RGB-D} benchmark,'' in
  \emph{Workshop on Color-Depth Camera Fusion in Robotics at the IEEE/RJS
  International Conference on Intelligent Robot Systems}, 2012.

\bibitem{KITTI}
A.~Geiger, P.~Lenz, C.~Stiller, and R.~Urtasun, ``Vision meets robotics: The
  {KITTI} dataset,'' \emph{The International Journal of Robotics Research},
  vol.~32, no.~11, pp. 1231--1237, 2013.

\bibitem{nardi2015introducing}
L.~Nardi, B.~Bodin, M.~Z. Zia, J.~Mawer, A.~Nisbet, P.~H. Kelly, A.~J. Davison,
  M.~Luj{\'a}n, M.~F. O'Boyle, G.~Riley \emph{et~al.}, ``Introducing
  {SLAMBench}, a performance and accuracy benchmarking methodology for
  {SLAM},'' in \emph{IEEE International Conference on Robotics and Automation},
  2015, pp. 5783--5790.

\bibitem{bodin2018slambench2}
B.~Bodin, H.~Wagstaff, S.~Saecdi, L.~Nardi, E.~Vespa, J.~Mawer, A.~Nisbet,
  M.~Luj{\'a}n, S.~Furber, A.~J. Davison \emph{et~al.}, ``{SLAMBench2}:
  Multi-objective head-to-head benchmarking for visual {SLAM},'' in \emph{IEEE
  International Conference on Robotics and Automation}, 2018, pp. 1--8.

\bibitem{saeedi2018navigating}
S.~Saeedi, B.~Bodin, H.~Wagstaff, A.~Nisbet, L.~Nardi, J.~Mawer, N.~Melot,
  O.~Palomar, E.~Vespa, T.~Spink \emph{et~al.}, ``Navigating the landscape for
  real-time localization and mapping for robotics and virtual and augmented
  reality,'' \emph{Proceedings of the IEEE}, no.~99, pp. 1--20, 2018.

\bibitem{wang2017stereo}
R.~Wang, M.~Schworer, and D.~Cremers, ``Stereo {DSO}: Large-scale direct sparse
  visual odometry with stereo cameras,'' in \emph{IEEE International Conference
  on Computer Vision}, 2017, pp. 3903--3911.

\bibitem{sun2018robust}
K.~Sun, K.~Mohta, B.~Pfrommer, M.~Watterson, S.~Liu, Y.~Mulgaonkar, C.~J.
  Taylor, and V.~Kumar, ``Robust stereo visual inertial odometry for fast
  autonomous flight,'' \emph{IEEE Robotics and Automation Letters}, vol.~3,
  no.~2, pp. 965--972, 2018.

\bibitem{delmerico2018benchmark}
J.~Delmerico and D.~Scaramuzza, ``A benchmark comparison of monocular
  visual-inertial odometry algorithms for flying robots,'' \emph{IEEE
  International Conference on Robotics and Automation}, vol.~10, p.~20, 2018.

\bibitem{qin2018vins}
T.~Qin, P.~Li, and S.~Shen, ``{VINS-Mono}: A robust and versatile monocular
  visual-inertial state estimator,'' \emph{IEEE Transactions on Robotics},
  vol.~34, no.~4, pp. 1004--1020, 2018.

\bibitem{zhao2019maphash}
Y.~Zhao, W.~Ye, and P.~Vela, ``Low-latency visual {SLAM} with
  appearance-enhanced local map building,'' in \emph{IEEE International
  Conference on Robotics and Automation}, 2019, pp. 8213--8219.

\end{thebibliography}

%

\begin{IEEEbiography}[{\includegraphics[width=1in,height=1.25in,clip,keepaspectratio]{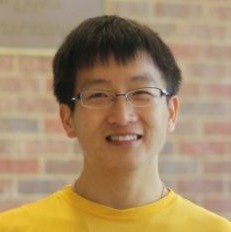}}]{Yipu Zhao}
obtained his Ph.D. in 2019, under the supervision of Patricio A. Vela, at the School of Electrical and Computer Engineering, 
Georgia Institute of Technology, USA.  Previously he received his B.Sc. degree in 2010 and M.Sc. degree in 2013,
at the Institute of Artificial Intelligence, Peking University, China.  His research interests include visual odometry/SLAM, 
3D reconstruction, and multi-object tracking.
\end{IEEEbiography}
\begin{IEEEbiography}[{\includegraphics[width=1in,height=1.25in,clip,keepaspectratio]{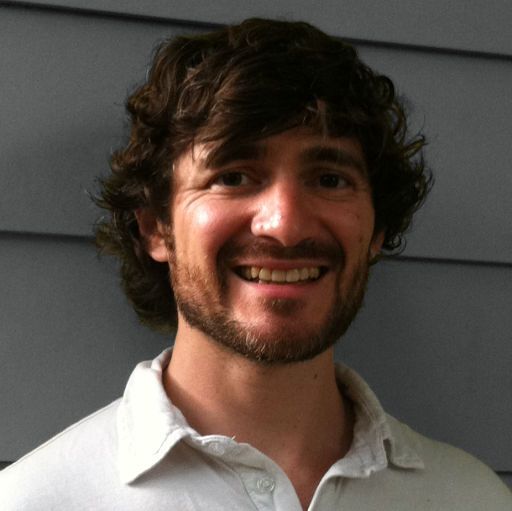}}]{Patricio A. Vela}
is an associate professor in the School of Electrical and Computer Engineering, and the Institute of Robotics and Intelligent Machines, at Georgia Institute of Technology, USA. His research interests lie in the geometric perspectives to control theory and computer vision. Recently, he has been interested in the role that computer vision can play for achieving control-theoretic objectives of (semi-)autonomous systems. His research also covers control of nonlinear systems, typically robotic systems.

Prof. Vela earned his B.Sc. degree in 1998 and his Ph.D. degree in control and dynamical systems in 2003, both from the California Institute of Technology, where he did his graduate research on geometric nonlinear control and robotics. In 2004, Dr. Vela was as a post-doctoral researcher on computer vision with School of ECE, Georgia Tech. He join the ECE faculty at Georgia Tech in 2005.

Prof. Vela is a member of IEEE.
\end{IEEEbiography}




\end{document}